\documentclass[11pt]{article}

\usepackage[preprint]{acl}
\usepackage{xcolor}
\usepackage{CJKutf8}
\usepackage{wrapfig}
\usepackage[table]{xcolor}
\usepackage{booktabs}    
\usepackage{xcolor}      
\usepackage{colortbl}    
\usepackage{graphicx} 

\definecolor{cvprblue}{rgb}{0.21,0.49,0.74}
\definecolor{Ocean}{RGB}{129,194,234}
\definecolor{Mygreen}{RGB}{170, 240, 238}
\definecolor{Magenta}{rgb}{0.8, 0.1, 0.6}

\usepackage{times}
\usepackage{latexsym}
\usepackage{multirow}

\usepackage[T1]{fontenc}
\usepackage[utf8]{inputenc}
\usepackage{microtype}
\usepackage{inconsolata}
\usepackage{graphicx}
\usepackage{enumitem}
\usepackage{booktabs}
\usepackage{makecell}

\title{Open Multimodal Retrieval-Augmented Factual Image Generation}

\usepackage[most]{tcolorbox}

\tcbset{
  promptbox/.style={
    colback=white,
    colframe=black,
    fonttitle=\bfseries,
    title=Prompt,
    boxrule=0.1mm,
    arc=1mm,
    outer arc=1mm,
    top=1mm,
    bottom=1mm,
    left=1mm,
    right=1mm
  }
}

\author{
Yang Tian\footnotemark[3]\thanks{Work done during an internship at Kuaishou Technology.}, Fan Liu\footnotemark[2], Jingyuan Zhang\footnotemark[6], Wei Bi\footnotemark[5], Yupeng Hu\footnotemark[3], Liqiang Nie\footnotemark[4] \\
\footnotemark[3] Shandong University, \footnotemark[2] National University of Singapore,\\
\footnotemark[6]Kuaishou Technology, \footnotemark[5] Independent Researcher\\
\footnotemark[4] Harbin Institute of Technology, Shenzhen\\
\small\texttt{\{tianyangchn,liufancs,nieliqiang\}@gmail.com}\\
}

\begin{document}
\maketitle
\begin{abstract}

Large Multimodal Models (LMMs) have achieved remarkable progress in generating photorealistic and prompt-aligned images, but they often produce outputs that contradict verifiable knowledge, especially when prompts involve fine-grained attributes or time-sensitive events. Conventional retrieval-augmented approaches attempt to address this issue by introducing external information, yet they are fundamentally incapable of grounding generation in accurate and evolving knowledge due to their reliance on static sources and shallow evidence integration. To bridge this gap, we introduce ORIG, an agentic open multimodal retrieval-augmented framework for Factual Image Generation (FIG), a new task that requires both visual realism and factual grounding. ORIG iteratively retrieves and filters multimodal evidence from the web and incrementally integrates the refined knowledge into enriched prompts to guide generation. To support systematic evaluation, we build FIG-Eval, a benchmark spanning ten categories across perceptual, compositional, and temporal dimensions. Experiments demonstrate that ORIG substantially improves factual consistency and overall image quality over strong baselines, highlighting the potential of open multimodal retrieval for factual image generation.
\end{abstract}

\section{Introduction}

\begin{figure}
\centering
\includegraphics[width=1\linewidth]{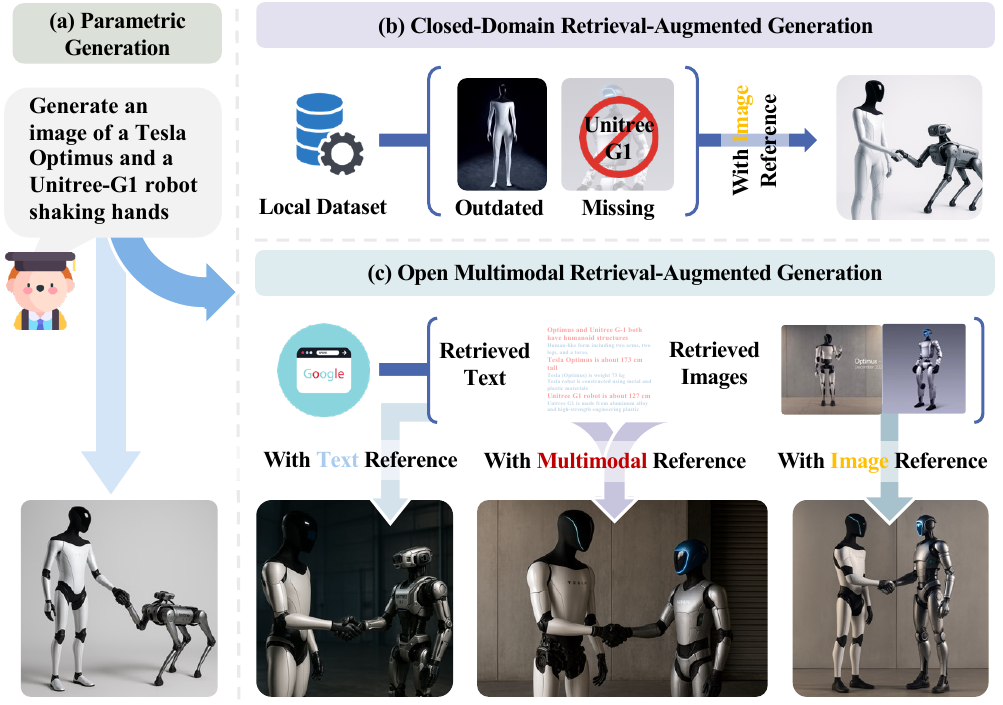}
\caption{Motivation of factual image generation (FIG) with open multimodal retrieval. (a) Reliance on internal knowledge alone often leads to outdated or hallucinated content. (b) Incorporating external information improves grounding but remains constrained by static and unimodal sources. (c) Leveraging open retrieval of multimodal evidence integrates evolving knowledge and complementary cues to achieve FIG.}
\label{fig:task}
\vspace{-0.3 cm}
\end{figure}

Recent advances in image generation have enabled models to produce photorealistic and compositionally rich visual content~\cite{karras2019style, rombach2022high, zhao2024monoformer, wu2025qwen}. Large multimodal models (LMMs) further extended these capabilities by jointly modeling language and vision under a unified framework~\cite{fuguiding,wu2025qwen, wang2025gpt}. They can flexibly interpret open-domain prompts, compose multiple entities, and integrate contextual signals to create visually compelling results.

Despite these advances, existing generation paradigms, including both standalone LMMs~\cite{bai2024hallucination,huang2025survey} and retrieval-augmented approaches~\cite{humrag,yuan2025finerag}, still struggle with factual consistency. Beyond creative applications, image generation plays an increasing role in education and scientific communication, where generated visuals are expected to convey accurate knowledge rather than merely appealing visuals. 
This issue becomes especially pronounced when prompts involve fine-grained attributes, such as relative scale or material composition, or require up-to-date knowledge about real-world entities and events.

As illustrated in Figure~\ref{fig:task}, generating “Tesla Optimus shaking hands with Unitree-G1” demands more than basic object recognition. Textual knowledge is needed to capture factual details such as relative heights and materials, while visual references provide essential cues about appearance, proportions, and spatial configuration. Moreover, prompts that reference evolving real-world events, such as new product releases or sports outcomes, require timely knowledge updates that are beyond the capabilities of both static pre-trained models and conventional retrieval-augmented approaches reliant on fixed corpora.

The limitation arises from two fundamental factors. First, current generation paradigms rely on static parametric memory or closed-domain corpora, making them unable to capture new facts, evolving entity states, or time-sensitive events. Second, factual grounding requires complementary multimodal evidence. Text provides attributes and relations, while images provide perceptual cues such as appearance, proportion, and spatial configuration. However, most existing approaches process these modalities separately and use retrieved knowledge only as auxiliary information, limiting their ability to generate factually grounded content.

Motivated by these challenges, we formalize \textbf{F}actual \textbf{I}mage \textbf{G}eneration (\textbf{FIG}) as a new task setting that requires outputs to be both visually realistic and factually grounded. 
To address this task, we propose \textbf{ORIG}, an agentic \textbf{O}pen multimodal \textbf{R}etrieval-augmented framework for \textbf{I}mage \textbf{G}eneration. The framework consists of three key components: an open multimodal retrieval module that iteratively gathers external evidence and performs coarse cross-modal filtering to discard irrelevant content; a prompt construction module that extracts fine-grained factual elements from the filtered evidence and assembles them into a generation-ready prompt; and an image generation module that produces the final output guided by this enriched prompt. 

To evaluate the effectiveness of multimodal knowledge integration, we further introduce \textbf{FIG-Eval} (\textbf{F}actual \textbf{I}mage \textbf{G}eneration \textbf{Eval}uation). FIG-Eval spans ten categories annotated across three key conceptual dimensions: perceptual elements, object and scene composition, and temporal sequences. It provides a comprehensive benchmark for evaluating models’ ability to leverage retrieved text and images for factual image generation. To facilitate reproducibility, code and data are released at \url{https://tyangjn.github.io/orig.github.io/}. 
In summary, our main contributions are as follows:
\begin{itemize}[leftmargin=*, nolistsep, noitemsep]
\item We formalize Factual Image Generation as a new task setting that emphasizes factual grounding in addition to realism.
\item We propose ORIG, an agentic open multimodal retrieval-augmented framework that iteratively retrieves evidence from the web and distills it into task-relevant knowledge for generation.
\item We construct FIG-Eval, a benchmark spanning ten entity classes and three concept dimensions, enabling systematic evaluation of factual consistency in image generation.
\item Extensive experiments demonstrate that ORIG consistently improves factual grounding and overall generation quality across diverse settings.
\end{itemize}

\section{Related Work}
\textbf{Advances in Image Generation.}
Image generation has progressed from GAN-based models \cite{zhang2017stackgan} to diffusion models \cite{rombach2022high, ramesh2022hierarchical, saharia2022photorealistic}, which significantly improved fidelity, diversity, and controllability. Recent developments have shifted toward LLM-based architectures, which can be broadly categorized into two primary paradigms: auto-regressive models that generate images from unified visual-text tokens \cite{wang2024emu3, batifol2025flux}, and hybrid models that use an LLM backbone to steer a diffusion decoder \cite{zhoutransfusion, wu2025qwen, geng2025x}. However, despite their impressive realism, these models often suffer from factual inconsistency. This issue primarily stems from a fundamental gap between the models' static, internal knowledge and the dynamic, real-world facts~\cite{kalai2024calibrated, xu2025mitigating}. This limitation undermines their reliability in critical domains such as education, media, and science, where accuracy is crucial~\cite{robertson2024google,liu2023riatig}.


\textbf{Retrieval-Augmented Generation.} 
To improve the factual grounding, recent works retrieve reference images from local datasets to guide generation~\cite{ shalev2025imagerag}. While this improves visual fidelity, methods such as FineRAG~\cite{yuan2025finerag} and Tiger~\cite{qutiger} are fundamentally limited by their reliance on closed-domain corpora with static image references, which fail to capture evolving real-world knowledge~\cite{hu2024instruct}. In contrast, existing multimodal web-retrieval frameworks (e.g., OmniSearch~\cite{li2024benchmarking}, OpenManus~\cite{openmanus2025}) are primarily designed for text generation. They often treat images as auxiliary signals, which limits their ability to filter noisy data and selectively exploit visual evidence, both of which are essential for generating complex and factually consistent images.

\textbf{Evaluation Benchmarks for Image Generation.}
Evaluation benchmarks for image generation have traditionally centered on metrics like prompt alignment, image quality, and compositional correctness~\cite{hu2024instruct, li2024evaluating, xu2023imagereward, li2024unimo, niu2025wise}. While recent efforts have expanded this scope to include more complex reasoning tasks, such as commonsense and physical understanding~\cite{pengdreambench++, huang2024t2i, fucommonsense, yan2025gpt}. However, a critical dimension remains largely underexplored. The factual accuracy of generated content, particularly in scenarios requiring the integration of multimodal external knowledge, is not adequately addressed by current evaluation protocols.

\section{Methodology}
\subsection{Task Definition}
We formalize FIG as a new task setting where the defining goal is to ensure factual consistency in generated images. Formally, given a query prompt $P$, the task requires producing an image that is semantically aligned with $P$, and grounded in verifiable knowledge about entities, attributes, relations, and temporal events. Specifically, factual consistency in FIG spans three dimensions: \textbf{Perceptual Fidelity}, ensuring faithful perception and accurate rendering of objects’ visual appearance; \textbf{Compositional Consistency}, which enforces accurate object properties and spatial relations, and \textbf{Temporal Consistency}, which ensures proper depiction of event timing and entity states. Unlike conventional image generation, FIG requires grounding in external evidence beyond the limited and static parametric memory of LMMs. In practice, this necessitates open retrieval from the web, where textual and visual evidence contribute complementary knowledge to support Perceptual Fidelity and Compositional Consistency, while the real-time nature of retrieval supplies updated information essential for Temporal Consistency.



\begin{figure*}
\centering
\includegraphics[width=0.9\textwidth]{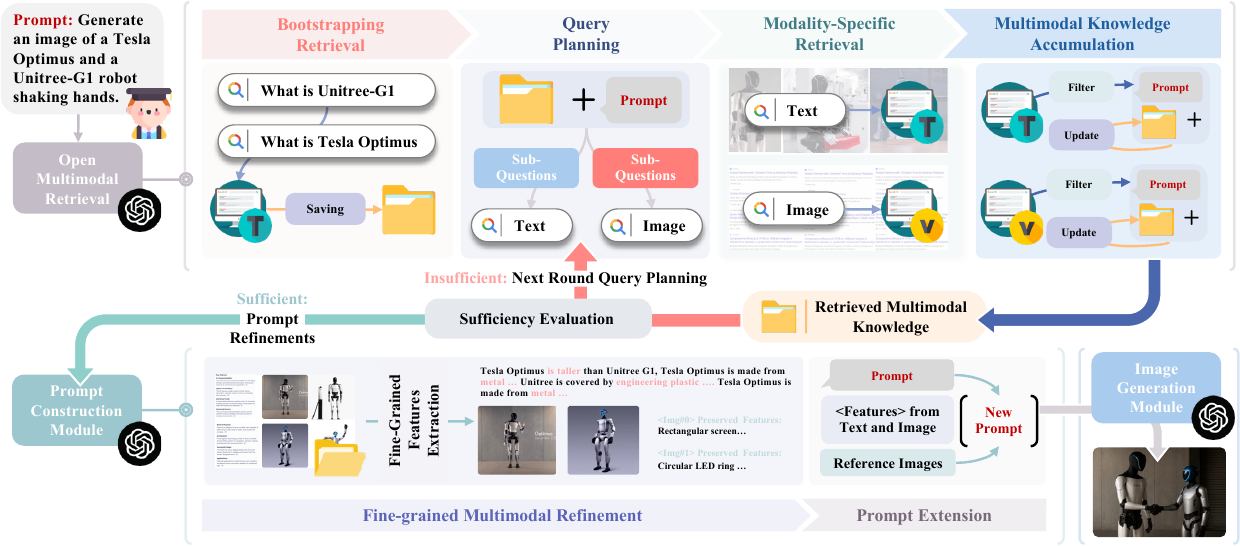}
\vspace{-0.2cm}
\caption{\textbf{The overall pipeline of the ORIG framework}. ORIG adaptively controls multimodal retrieval and prompt construction, dynamically deciding whether to continue retrieval or proceed based on the current state of accumulated knowledge.}
\vspace{-0.6cm}
\label{fig:pipeline}
\end{figure*}

\subsection{The ORIG Framework}
We propose ORIG, an agentic open multimodal retrieval-augmented framework that grounds image generation in verifiable knowledge. As illustrated in Figure~\ref{fig:pipeline}, ORIG adopts an iterative pipeline that plans sub-queries, retrieves modality-specific evidence, filters noise through coarse-to-fine filtering, and integrates refined knowledge into enriched prompts that guide factual image generation. The framework comprises three modules: \textbf{Open Multimodal Retrieval Module} that collects and filters web-scale evidence through adaptive query planning and sufficiency evaluation; \textbf{Prompt Construction Module} that integrates the input prompt with extracted features from filtered evidence to create generation-ready prompts, and \textbf{Image Generation Module} that produces factually grounded images based on the enriched prompts.

\subsubsection{Open Multimodal Retrieval}
This module is implemented as an agentic open-retrieval loop that incrementally builds a reliable knowledge base from the web. ORIG operates as a closed feedback cycle guided by both the input prompt and the accumulated external knowledge. The loop consists of five stages:

\textbf{Bootstrapping Retrieval.} To ensure effective query planning, the module begins with a lightweight bootstrapping retrieval before entering the retrieval loop, which provides basic knowledge of the entities or concepts contained in the prompt $P$, preventing the model from generating misaligned sub-queries due to insufficient understanding of rare or novel terms. The retrieval results are stored in the local knowledge base $\mathbf{K}$, serving as the initial context for subsequent query-planning and enabling early error correction.

\textbf{Query Planning.} The stage begins by analyzing the input $P$ against its current $\mathbf{K}$ to identify under-specified or missing information. These identified knowledge gaps are then decomposed into a set of sub-questions $\mathbf{Q}$, which are mapped to their optimal modality. This complementary mapping generates textual queries $\mathbf{S}^t$ to retrieve contextual knowledge (e.g., attributes, relations), while visual queries $\mathbf{S}^v$ are designed to capture perceptual information (e.g., appearance, spatial configurations). Instructed by $I_{\text{Q}}$, the output of retrieval backbone model $\mathcal{M}$~\footnote{We denote the large multimodal model as $\mathcal{M}$, which serves as the core reasoning and generation component. In our implementation, $\mathcal{M}$ is instantiated with models such as Qwen-VL and GPT-5, while the framework itself remains model-agnostic.
} for planning stage is: 
\vspace{-0.1cm}
\begin{equation}
    \small
 \langle\mathbf{Q},\mathbf{S}^t,\mathbf{S}^v\rangle =\mathcal{M}(I_{\text{Q}},P, \mathbf{K}).
    \vspace{-0.1cm}
\end{equation}

\textbf{Modality-Specific Retrieval.} The module performs retrieval using publicly available web retrieval APIs, one for each modality. The outputs are denoted as $\mathbf{R}^t$ and $\mathbf{R}^v$, which correspond to the raw retrieved text and image sets, respectively. For each query $S$ in the query sets $\mathbf{S}^t$ and $\mathbf{S}^v$, the appropriate API is invoked:
\vspace{-0.1cm}
\begin{equation}
    \small
    \mathbf{R}^m = \{ \text{API}^m(S) \mid S \in \mathbf{S}^m \}, \quad m \in \{t, v\}. 
    \vspace{-0.1cm}
\end{equation}

\textbf{Multimodal Knowledge Accumulation.} The module incrementally builds a relevant knowledge base by performing coarse-grained filtering on all retrieved content. The filtering operates in two steps: First, the model evaluates retrieved text snippets $\mathbf{R}^t$ against the existing multimodal context. A text is retained only if it satisfies the joint conditions of semantic alignment with the prompt and factual consistency with the multimodal knowledge base $\textbf{K}$:
\vspace{-0.1cm}
\begin{equation}
\small
\hat{\mathbf{R}}^t = \mathcal{M}(I_{\text{TF}}, P, \mathbf{K}, \mathbf{R}^t),
\vspace{-0.05cm}
\end{equation}
where $I_{\text{TF}}$ guides the textual filtering process. Next, the module applies similar multimodal consistency-based filtering to retrieved images $\mathbf{R}^v$, retaining those that maintain coherence with both textual evidence and existing visual evidence:
\vspace{-0.1cm}
\begin{equation}
\small
\hat{\mathbf{R}}^v = \mathcal{M}(I_{\text{VF}}, P, \mathbf{K}, \mathbf{R}^v),
\vspace{-0.1cm}
\end{equation}
where $I_{\text{IF}}$ is the instruction for image filtering. The filtered texts $\hat{\mathbf{R}}^t$ and images $\hat{\mathbf{R}}^v$ are incorporated into the knowledge base, which is updated for subsequent iterations:
\vspace{-0.1cm}
\begin{equation}
\small
\mathbf{K} \leftarrow \mathbf{K} \cup \{ \hat{\mathbf{R}}^t, \hat{\mathbf{R}}^v \}.
\vspace{-0.1cm}
\end{equation}

\begin{table*}[t]
    \centering
    \caption{Prompt and Question Distribution Across 10 Entity Classes and three Concept Categories. The entity classes include Animal (An.), Sports (Sp.), Transportation (Tr.), Landmarks (La.), Food (Fo.), People (Pe.), Plants (Pl.), Products (Pr.), Culture (Cu.), and Events (Ev.).}
    \vspace{-0.3cm}
    \small
    \scalebox{0.85}{
    \begin{tabular}{l|rrrrrrrrrr|r}
    \toprule
    \textbf{Categories} & \textbf{An.} & \textbf{Sp.} & \textbf{Tr.} & \textbf{La.} & \textbf{Fo.} & \textbf{Pe.} & \textbf{Pl.} & \textbf{Pr.} & \textbf{Cu.} & \textbf{Ev.} & \textbf{Total} \\
    \midrule
    \rowcolor{lightgray!30}
    Prompt Number                              & 55  & 52  & 50   & 50   & 49  & 52   & 51  & 56  & 50  & 49  & 514 \\
    Perceptual Fidelity (\textbf{PF})          & 233 & 197 & 207  & 138  & 116 & 194  & 219 & 287 & 171 & 105 & 1,867 \\
    Compositional Consistency (\textbf{CC})    & 120 & 148 & 143  & 148  & 189 & 185  & 117 & 106 & 191 & 243 & 1,590 \\
    Temporal Consistency (\textbf{TC})         & 73  & 46  & 44   & 95   & 88  & 40   & 89  & 31  & 65  & 65  & 636 \\
    \rowcolor{lightgray!30}
    All Concepts Categories                    & 426 & 391 & 394  & 381  & 388 & 419 & 425  & 418 & 427 & 413 & 4,093 \\
    \bottomrule
    \end{tabular}}
    \vspace{-0.3cm}
    \label{tab:entity_categories_transposed}
\end{table*}

\textbf{Sufficiency Evaluation.} Once the knowledge base is updated, the module determines whether the retrieved knowledge sufficiently addresses the identified sub-queries $\mathbf{Q}$. This is formalized as:
\vspace{-0.1cm}
\begin{equation}
    \small
    D = \mathcal{M}(I_{\text{SE}}, P, \mathbf{K}, \mathbf{Q}), D\in\{\text{Retrieval}, \text{Refine}\}
    \vspace{-0.1cm}
\end{equation}
where $I_{\text{SE}}$ is the instruction guiding sufficiency evaluation. If the knowledge base is incomplete or lacks essential details, the module is advised to initiate another retrieval round. Otherwise, it proceeds to the next stage of prompt construction. This feedback-driven control mechanism enables ORIG to adaptively determine the optimal number of retrieval rounds required to support grounded and informed image generation.

\subsubsection{Prompt Construction and Image Generation}
Once the retrieval module determines that sufficient evidence has been gathered, ORIG transitions to the prompt construction module. This module transforms the multimodal knowledge base $\textbf{K}$ into a generation-ready prompt through fine-grained knowledge refinement and prompt extension. The enhanced prompt is then fed into the image generation module for factually grounded synthesis.

\textbf{Fine-grained Multimodal Refinement.} While coarse-grained filtering during retrieval eliminated obviously irrelevant content, fine-grained refinement focuses on precision and semantic quality. This stage applies stricter criteria to extract the most discriminative features from the remaining multimodal evidence, moving beyond simple relevance to identify content that directly supports generation objectives. The refinement process operates in two stages. First, the module identifies visual descriptors and generation-relevant attributes within the textual knowledge to yield a set of core textual features $\mathbf{F}^t$, while simultaneously deduplicating the image set $\mathbf{K}^v$ to refined image set $\mathbf{K}^v_R$:
\begin{equation}
\vspace{-0.1cm}
\small
\langle \mathbf{F}^t, \mathbf{K}^v_R \rangle = \mathcal{M}(I_{\text{CR}}, P, \mathbf{K}),
\vspace{-0.1cm}
\end{equation}
where $I_{\text{CR}}$ guides the extraction of visually-descriptive textual features and image deduplication. Subsequently, visual control features are extracted from $\mathbf{K}^v_R$ using cross-modal guidance:
\vspace{-0.1cm}
\begin{equation}
    \small
    \mathbf{F}^v = \mathcal{M}(I_{\text{VR}}, P, \mathbf{K}^v_R, \mathbf{F}^t),
    \vspace{-0.1cm}
\end{equation}
where $I_{\text{VR}}$ leverages textual features $\mathbf{F}^t$ to identify critical visual elements that should be preserved for generation control, while filtering out irrelevant background elements that could compromise generation fidelity. This process produces attention guidance that directs the generator to focus on essential visual cues within the reference images.

\textbf{Prompt Extension.} The extracted multimodal features are integrated to synthesize the final generation prompt:
\vspace{-0.1cm}
\begin{equation}
    \small
    \hat{P} = \mathcal{M}(I_{\text{PE}}, P, \mathbf{F}^v, \mathbf{K}^v_R, \mathbf{F}^t),
    \vspace{-0.1cm}
\end{equation}
where $I_{\text{PE}}$ guides the extension process. This enriched textual prompt $\hat{P}$ not only incorporates the retrieved factual knowledge but also instructs the model to focus on critical visual elements within the reference images. The prompt $\hat{P}$ and filtered reference images $\mathbf{K}^v_R$ are jointly fed to an image generation module for multimodal factually grounded synthesis.

\subsection{Discussion}
ORIG can be viewed as a self-adaptive retrieval framework that rethinks how external knowledge is integrated into image generation. Traditional unimodal retrieval approaches often rely on static visual references, which limit their ability to incorporate abstract concepts, contextual cues, and time-sensitive information. In contrast, ORIG employs an iterative agentic retrieval process that plans sub-queries, filters relevant evidence, and incrementally refines a multimodal knowledge base conditioned on the evolving prompt state. This dynamic retrieval–integration cycle enables the model to go beyond static grounding, mitigating factual inconsistencies and aligning generated images with evolving real-world knowledge.

\section{Dataset: FIG-Eval}
We introduce FIG-Eval, a curated dataset designed to evaluate whether image-generation models can effectively leverage web-retrieved multimodal evidence to achieve FIG. It covers 10 entity classes and three concept groups, featuring knowledge-intensive prompts that encode implicit, domain-specific facts requiring external evidence beyond parametric knowledge. For reliable assessment, each prompt is paired with human-annotated ground-truth references. From these evidence, we manually derive multimodal QA questions that rigorously operationalize factual correctness, enabling automated scoring via state-of-the-art Vision-Language Models.

\textbf{Prompt Construction.} Building on prior typologies~\cite{lim2025evaluating,niu2025wise,huang2024t2i}, we established a taxonomy of ten entity classes (Table~\ref{tab:entity_categories_transposed}) and constructed a retrieval-oriented dataset guided by two key principles: multi-hop sequential retrieval, which enables stepwise text-to-image grounding, and parallel co-retrieval, which jointly integrates textual and visual evidence. Subsequently, five expert annotators (each with over one year of evaluation experience) curated 1,132 prompts, each paired with $\sim$8 annotated multimodal references. To ensure retrieval dependence, we applied a two-stage filtering process. First, ambiguous, underspecified, or trivial items were removed. Second, the remaining prompts were stress-tested with GPT, Gemini, and Qwen2.5-VL under no-retrieval settings, instructing models direct generation with parametric knowledge. If the generated images faithfully reflected all intended reference facts, the prompts were deemed solvable without retrieval and discarded. This process yielded 514 validated prompts, balanced across 10 entity classes.


\begin{table}[t]
    \centering
    \small
    \vspace{-0cm}
    \caption{Visual concepts by categories.}
    \vspace{-0.3cm}
    \scalebox{0.83}{
    \begin{tabular}{l|l}
    \toprule
    \textbf{Concepts Categories}       & \textbf{Visual Concepts} \\
    \midrule
    Perceptual Fidelity (\textbf{PF})                 & Color, Appearance, Size, Texture  \\
    \midrule
    Compositional                                     & Object Number, Position, Interaction, \\
    Consistency (\textbf{CC})                         & Fore-Background \\
    \midrule
    Temporal Consistency (\textbf{TC})                & Time, Sequence Order, Process Steps\\
    \bottomrule
    \end{tabular}}
     \vspace{-0.3cm}
    \label{tab:visual_concepts}
\end{table}

\textbf{Evaluation.} 
Following prior dataset designs \cite{niu2025wise,huang2025t2i}, we group factual information into three visual concept categories (Table~\ref{tab:visual_concepts}): Perceptual Fidelity, Compositional Consistency, and Temporal Consistency.  
Each prompt $P_i$ is paired with expert-annotated references from which we extract a ground-truth feature set  
$\mathbf{F}_i^{\text{GT}}=\{F_{ij}^{\text{GT}}\}_{j=1}^{n_i}$ ($n_i=|\mathbf{F}_i^{\text{GT}}|$).  
Annotators then construct multimodal true–false questions  
$\mathbf{X}_i=\{X_{ik}\}_{k=1}^{m_i}$ ($m_i=|\mathbf{X}_i|$), each mapped to a concept group, with all answers fixed as \textbf{True}.  
This yields 4,093 questions in total, all dual-annotated with adjudication. Generated images $I_i$ are then evaluated with model $\hat{\mathcal{M}}$ (GPT-5) under a fixed QA template. For each question $X_{ik}$, the model outputs score $Y_{ik}\!\in\!\{\text{True},\text{False}\}$:
\vspace{-0.1cm}
\begin{equation}
    \small
    Y_{ik} = 
    \begin{cases} 
    \mathcal{\hat{M}}(X_{ik}, I_i, I^{\text{GT}}) & \text{required reference image} \\
    \mathcal{\hat{M}}(X_{ik}, I_i), & \text{else}
    \end{cases}
\end{equation}
The per-prompt accuracy is the mean of $\{Y_{ik}\}_{k=1}^{m_i}$, which is then macro-averaged for concept- and class-level reporting:
\vspace{-0.1cm}
\begin{equation}
\small
S_i = \frac{1}{m_i}\sum_{k=1}^{m_i}\mathbf{1}[Y_{ik}=\text{True}].
\vspace{-0.1cm}
\end{equation}
Given retrieved text $\mathbf{R}_i^t$ and images $\mathbf{R}_i^v$, we also evaluate retrieval quality by measuring the alignment between $\{\mathbf{R}_i^t, \mathbf{R}_i^v\}$ and $\mathbf{F}_{i}^{\text{GT}}$ with $\hat{\mathcal{M}}$:
\vspace{-0.1cm}
\begin{equation}
\small
A_i^{\text{ret}} = \frac{1}{n_i}\sum_{j=1}^{n_i}\mathrm{Align}(F_{ij}^{\text{GT}}\mid \mathbf{R}_i^t, \mathbf{R}_i^v),
\vspace{-0.1cm}
\end{equation}
where $\mathrm{Align}\!\in\!\{0,1\}$. This yields a feature-level retrieval accuracy in $[0,1]$.

\textbf{Correlation with Human Judgments.} To validate our automated evaluator $\hat{\mathcal{M}}$, we compared it against a human baseline. We randomly sampled 200 prompts and generated images using three image generation models (GPT-Image, Gemini-Image, and Qwen-Image). Five expert annotators independently answered the QA pairs for all model outputs, with majority voting used for final labels. We observe strong agreement between evaluator and human scores, with correlations computed over per-class averages (Pearson’s $r=0.929$, Spearman’s $\rho=0.936$, Kendall’s $\tau=0.772$).  To further test robustness and mitigate potential bias, we evaluated on Gemini-2.5-Flash outputs, which again yielded high consistency ($r=0.922$, $\rho=0.928$, $\tau=0.764$), confirming the reliability and generalizability of our protocol.

\section{Experiment}
\subsection{Experimental Setup}
We evaluate on FIG-Eval by jointly considering image generators and retrieval agents. For generation, we select three representative LLM-based image generators with strong multimodal reasoning capacity: GPT-Image (GPT-Image-1), Gemini-Image (Gemini-2.5-Flash-Image), and the open-source Qwen-Image. Each generator is tested under four knowledge conditions: \textbf{Direct}, using only the raw prompt; \textbf{Oracle}, conditioned on annotated gold textual and visual references as an upper bound; \textbf{Prompt Enhanced}, where prompts are expanded with parametric knowledge from the retrieval backbone LLM; and \textbf{Retrieval}, where prompts are augmented with retrieved evidence. Within the retrieval setting, we compare our ORIG agent against two open-source multimodal baselines, OpenManus~\citep{openmanus2025} and OmniSearch~\citep{li2024benchmarking}, and additionally introduce two modality-restricted variants: ORIG-Img, which retrieves only images, and ORIG-Txt, which retrieves only text. The retrieval agents are instantiated with two backbone LLMs, GPT-5 and Qwen2.5-VL-72B. Open retrieval is conducted via Google Search and Google Image Search with identical configurations across methods. All evaluation results are reported using GPT-5 as the evaluator to maintain consistency across settings.

\begin{table*}[!ht]
\centering
\caption{Accuracy (\%) comparison across 10 entity classes and 3 concept categories on the FIG-Eval dataset using GPT-5 as the retrieval backbone. The \textbf{All Acc.} columns report average score of all 10 categories with both GPT-5 and Qwen2.5-VL-72B backbones.}
\label{tab:main_results}
\vspace{-0.3cm}
\small
\scalebox{0.77}{
\begin{tabular}{l|cccccccccc|ccc|cc}
\toprule
\multicolumn{1}{c}{\multirow{2}{*}{\textbf{Methods}}} 
& \multicolumn{10}{c}{\textbf{Entity Classes}} 
& \multicolumn{3}{c}{\textbf{Concepts}} 
& \multicolumn{2}{c}{\textbf{All Acc.}} \\
\cmidrule(lr){2-11} \cmidrule(lr){12-14} \cmidrule(lr){15-16}
\multicolumn{1}{c}{\textbf{}}
& \multicolumn{1}{c}{\textbf{An.}} 
& \multicolumn{1}{c}{\textbf{Sp.}} 
& \multicolumn{1}{c}{\textbf{Tr.}} 
& \multicolumn{1}{c}{\textbf{La.}} 
& \multicolumn{1}{c}{\textbf{Fo.}} 
& \multicolumn{1}{c}{\textbf{Pe.}} 
& \multicolumn{1}{c}{\textbf{Pl.}} 
& \multicolumn{1}{c}{\textbf{Pr.}} 
& \multicolumn{1}{c}{\textbf{Cu.}} 
& \multicolumn{1}{c}{\textbf{Ev.}} 
& \multicolumn{1}{c}{\textbf{PF}} 
& \multicolumn{1}{c}{\textbf{CC}} 
& \multicolumn{1}{c}{\textbf{TC}} 
& \multicolumn{1}{c}{\textbf{GPT}} 
& \multicolumn{1}{c}{\textbf{Qwen}} \\
\midrule
 Qwen-Image                & 16.4 & 9.3 & 19.3 & 31.3 & 12.8 & 16.7 & 23.3 & 12.3 & 30.2 & 17.7 & 21.5 & 18.2 & 14.0 & \multicolumn{2}{c}{19.0} \\
  \rowcolor{lightgray!30}
 \hspace{0.5em}Oracle      & 78.2 & 69.1 & 73.2 & 81.7 & 68.9 & 73.4 & 79.2 & 70.1 & 77.6 & 73.4 & 74.4 & 73.1 & 71.9 & \multicolumn{2}{c}{74.5}\\
 \hspace{0.5em}Prompt Enhanced   & 31.9 & 16.7 & 37.9 & 41.2 & 23.3 & 26.4 & 45.3 & 25.0 & 34.5 & 31.9 & 30.7 & 33.4 & 30.3 & 31.5 & 28.1\\
 \hspace{0.5em}OpenManus   & 37.1 & 20.6 & 42.3 & 43.1 & 26.5 & 26.7 & 45.4 & 35.2 & 42.9 & 33.5 & 36.2 & 34.4 & 35.5 & 35.4 & 32.2\\
 \hspace{0.5em}OmniSearch  & 35.4 & 19.1 & 41.8 & 42.4 & 26.4 & 26.1 & 44.8 & 34.5 & 41.9 & 33.1 & 35.6 & 34.6 & 32.3 & 34.7 & 31.3\\
 \rowcolor{Ocean!20}
 \hspace{0.5em}ORIG        & 39.0 & 25.8 & 45.2 & 46.5 & 32.5 & 29.7 & 52.4 & 40.2 & 45.5 & 36.7 & 41.9 & 38.5 & 37.2 & 39.7 & 36.1\\
 \rowcolor{Ocean!20}
 \hspace{0.5em}ORIG-Img    & 32.3 & 21.6 & 42.1 & 44.5 & 25.6 & 27.5 & 47.4 & 38.0 & 37.8 & 34.4 & 39.2 & 34.3 & 27.9 & 35.4 & 32.0\\
 \rowcolor{Ocean!20}
 \hspace{0.5em}ORIG-Txt    & 37.4 & 21.7 & 41.0 & 42.8 & 28.0 & 27.3 & 50.8 & 33.9 & 43.3 & 34.9 & 36.1 & 37.1 & 36.4 & 36.5 & 32.9\\
 \midrule
 Gemini-Image              & 47.0 & 20.0 & 32.3 & 45.2 & 31.0 & 22.3 & 51.3 & 22.6 & 42.5 & 30.4 & 34.9 & 34.4 & 35.1 & \multicolumn{2}{c}{34.6}\\
 \rowcolor{lightgray!30}
 \hspace{0.5em}Oracle      & 85.1 & 77.9 & 78.1 & 86.3 & 78.2 & 77.4 & 87.8 & 78.5 & 85.4 & 82.5 & 84.4 & 79.8 & 78.9 & \multicolumn{2}{c}{81.8}\\
 \hspace{0.5em}Prompt Enhanced   & 49.6 & 25.2 & 41.0 & 49.2 & 38.2 & 30.7 & 61.0 & 28.7 & 50.7 & 38.1 & 39.8 & 42.4 & 45.3 & 41.4 & 34.9\\
 \hspace{0.5em}OpenManus   & 50.9 & 33.1 & 46.4 & 51.2 & 33.9 & 35.5 & 62.1 & 41.3 & 46.4 & 42.5 & 46.2 & 42.6 & 43.9 & 44.4 & 36.1\\
 \hspace{0.5em}OmniSearch  & 49.5 & 31.6 & 44.8 & 50.4 & 33.4 & 34.2 & 61.5 & 38.4 & 45.7 & 42.1 & 44.0 & 43.1 & 41.6 & 43.3 & 35.3\\
 \rowcolor{Ocean!20}
 \hspace{0.5em}ORIG        & 54.0 & 41.3 & 53.6 & 57.3 & 42.5 & 42.0 & 68.7 & 51.1 & 50.5 & 49.4 & 52.4 & 50.0 & 53.9 & 51.4 & 41.6\\
 \rowcolor{Ocean!20}
 \hspace{0.5em}ORIG-Img    & 50.0 & 33.5 & 52.1 & 50.5 & 39.1 & 35.9 & 62.1 & 46.3 & 46.9 & 41.9 & 49.5 & 43.9 & 43.6 & 46.2 & 37.3\\
 \rowcolor{Ocean!20}
 \hspace{0.5em}ORIG-Txt    & 52.9 & 36.5 & 46.1 & 53.9 & 39.8 & 35.0 & 64.8 & 42.4 & 47.7 & 48.1 & 46.2 & 48.1 & 47.2 & 46.9 & 37.8\\
                            
 \midrule
 GPT-Image                 & 39.7 & 17.1 & 33.1 & 44.5 & 21.8 & 20.2 & 45.7 & 31.9 & 35.1 & 30.3 & 34.6 & 30.7 & 29.1 & \multicolumn{2}{c}{32.1}\\
 \rowcolor{lightgray!30}
 \hspace{0.5em}Oracle      & 83.3 & 75.2 & 78.3 & 85.2 & 74.6 & 76.7 & 87.2 & 79.6 & 85.1 & 82.1 & 83.1 & 79.0 & 78.3 & \multicolumn{2}{c}{80.8}\\
 \hspace{0.5em}Prompt Enhanced   & 48.1 & 27.4 & 41.7 & 47.2 & 36.1 & 30.0 & 59.8 & 26.5 & 45.5 & 39.5 & 39.5 & 40.7 & 43.5 & 40.4 & 32.5\\
 \hspace{0.5em}OpenManus   & 50.5 & 32.4 & 45.8 & 50.1 & 33.2 & 34.7 & 60.8 & 39.9 & 45.1 & 42.2 & 45.1 & 42.0 & 43.1 & 43.6 & 35.5\\
 \hspace{0.5em}OmniSearch  & 49.1 & 30.7 & 43.6 & 49.2 & 32.8 & 33.6 & 60.4 & 37.9 & 44.2 & 41.0 & 43.2 & 41.9 & 40.8 & 42.4 & 34.7\\
 \rowcolor{Ocean!20}
 \hspace{0.5em}ORIG        & 53.0 & 40.2 & 51.6 & 56.4 & 40.4 & 41.0 & 67.0 & 50.5 & 49.5 & 47.6 & 51.5 & 48.9 & 50.6 & 50.1 & 40.5\\
 \rowcolor{Ocean!20}
 \hspace{0.5em}ORIG-Img    & 49.3 & 32.5 & 49.3 & 49.0 & 37.7 & 33.9 & 61.6 & 44.8 & 46.2 & 40.6 & 48.8 & 42.3 & 41.0 & 44.9 & 36.1\\
 \rowcolor{Ocean!20}
 \hspace{0.5em}ORIG-Txt    & 52.2 & 35.0 & 44.6 & 53.3 & 37.8 & 34.1 & 63.6 & 40.6 & 47.7 & 45.4 & 44.5 & 47.6 & 46.4 & 45.8 & 37.2\\
\bottomrule
\end{tabular}}
\vspace{-0.3cm}
\end{table*}

\subsection{Main Results}
As shown in Table~\ref{tab:main_results}, the \textbf{Direct} setting establishes the baseline: Gemini-Image (34.6\%) and GPT-Image (32.1\%) outperform Qwen-Image (19.0\%), reflecting inherent differences in their parametric knowledge. Performance further improves in the \textbf{Prompt Enhanced} setting (e.g., GPT-Image reaches 40.4\%) by leveraging the retrieval backbone model’s reasoning ability; however, these gains remain limited, as the enhanced prompts depend solely on internal knowledge and fail to incorporate the detailed, up-to-date information provided by web-retrieval methods. In contrast, when augmented with our proposed \textbf{ORIG} framework, all three image generation models achieve the highest accuracies, with GPT-Image at 50.1\%, Gemini-Image at 51.4\%, and Qwen-Image at 39.7\%, consistently surpassing retrieval baselines such as OmniSearch and OpenManus. Moreover, the ORIG model outperforms both unimodal variants, confirming that joint multimodal retrieval effectively combines complementary visual and textual knowledge for superior performance. Nonetheless, even with \textbf{Oracle} knowledge, the models still struggle to capture fine-grained visual attributes, underscoring a persistent challenge in integrating multimodal information for faithful image synthesis.

\textbf{Impact of retrieval backbones.} The results reveal two distinct stages of improvement. The transition from the \textbf{Direct} to the \textbf{Prompt Enhanced} setting reflects each model’s ability to exploit internal parametric knowledge through reasoning-based prompt enrichment, whereas the subsequent gain from \textbf{Prompt Enhanced} to \textbf{Retrieval} captures its capacity to integrate external knowledge. Qwen2.5-VL-72B shows a modest average 3.3\% improvement in the first phase but a more substantial 7.6\% gain with retrieval. In contrast, GPT-5 achieves stronger internal reasoning gains (+9.2\%) and comparable retrieval gains (+9.3\%), demonstrating a more balanced ability to leverage both parametric and external knowledge.

\textbf{Performance on different concept categories.} Across the three generation models, the Temporal Consistency category yielded the largest average improvement (+21.2\%), underscoring the critical role of retrieval in supplying the temporal and dynamic knowledge that is often underrepresented in parametric memory. Interestingly, Perceptual Fidelity showed a slightly higher gain (+18.2\%) than the more structurally complex Compositional Consistency category (+17.9\%). This suggests that while compositional reasoning remains a known challenge, integrating knowledge for multi-entity scenes is inherently more complex due to the difficulty of aligning diverse contextual cues. Conversely, retrieval of fine-grained perceptual attributes provides clearer and more targeted information that models can integrate.

\textbf{Performance across different entity classes.} Based on knowledge update frequency~\cite{chen2021dataset}, entities are categorized into \textit{dynamic} classes (e.g., Transportation, Events, Sports) and \textit{stable} classes (e.g., Animals, Landmarks, Culture). On average improvement of ORIG across three generation models, dynamic categories show lower baseline accuracy but receive a substantial boost from retrieval, with large improvements observed in Products (+24.9\%) and Sports (+20.3\%). In contrast, stable categories exhibit stronger baseline performance and therefore experience smaller retrieval gains, such as in Animals (+14.3\%) and Culture (+12.5\%).

\subsection{Ablation Study}\label{sec:ab_study}
\textbf{Contribution of Different Modality.} 
As shown in Table~\ref{tab:main_results}, based on averages across the three generation models relative to the Direct baseline, ORIG-Img yields the largest gains on Perception Fidelity (PF, +15.4\%) but smaller gains on Compositional Consistency (CC, +12.4\%) and Temporal Consistency (TC, +11.4\%), reflecting weaker semantic constraints and harder alignment in multi-entity or cross-stage scenes (e.g., Events), where insufficient prompt control can steer the model toward irrelevant elements within reference images; in such cases, image-only retrieval can even trail the Prompt-Enhanced results on CC and TC. By contrast, ORIG-Txt delivers steadier improvements on CC (+16.7\%) and TC (+15.6\%) by supplying attributes, relations, and procedural cues, though it is less effective on PF (+12.3\%) due to missing fine-grained appearance detail. At the entity level, image retrieval provides larger marginal benefits for visually fast-evolving classes (Transportation +19.6\%; People +20.7\%), whereas text retrieval benefits appearance-stable classes (Animals +13.1\%; Plants +19.6\%). Finally, by jointly leveraging text for semantic constraints and images for appearance cues, and by enforcing coarse-to-fine filtering with prompt refinement, the ORIG framework mitigates insufficient semantic control and visual misalignment, yielding consistent gains across 10 entity classes and 3 concept categories.

\begin{table}[t]
\centering
\small
\caption{Retrieval performance comparison(\%) based on Gemini-Image, with "Acc" as retrieval accuracy, "Token" for average token count in generated prompts, "Image Num" for the average number of images per prompt. "Iters" represent the total number of retrieval iterations, including the bootstrapping retrieval, +$n$-Round represents $n$ iterations in the retrieval loop.}
\vspace{-0.3cm}
\scalebox{0.9}{
\begin{tabular}{l|ccccc}
\toprule
\textbf{Methods} & \textbf{Iters} & \textbf{R. Acc} & \textbf{Tokens} & \textbf{Images} & \textbf{G. Acc} \\
\midrule
OpenManus  & 2.3 & 63.4 & 483.3 & 1.3 & 44.4 \\
OmniSearch & 2.5 & 60.8 & 472.9 & 1.5 & 43.3 \\ 
\midrule
ORIG       & 2.8 & 74.7 & 519.1 & 2.1 & 51.4 \\
+1-Round   & 2.0 & 70.6 & 453.5 & 1.6 & 50.3 \\
+2-Round   & 3.0 & 75.0 & 536.4 & 2.3 & 51.4 \\
+3-Round   & 4.0 & 75.1 & 597.2 & 2.8 & 50.9 \\
\bottomrule
\end{tabular}}
\vspace{-0.65cm}
\label{tab:retrieval_methods}
\end{table}

\textbf{Comparison of Multimodal Retrieval Methods.} 
As shown in Table~\ref{tab:retrieval_methods}, our adaptive ORIG framework achieves a high retrieval accuracy of 74.7\%, clearly outperforming baseline methods such as OpenManus (63.4\%) and OmniSearch (60.8\%). Further analysis shows that ORIG constructs richer prompts (averaging 519.1 tokens) and leverages more reference images (averaging 2.1), indicating its stronger ability to capture and synthesize relevant and detailed multimodal information, which in turn enhances the final generation quality. To further examine retrieval efficiency, we conducted a study on fixed-round strategies in the retrieval loop. The results show that increasing the retrieval rounds to three only slightly improves retrieval accuracy to 75.1\%, while it leads to a small decrease in generation accuracy to 50.9\%. This trade-off likely arises from two factors: first, excessive retrieval rounds tend to introduce overly fine-grained or redundant descriptions of minor details that are difficult for the image generator to reflect effectively; second, a larger set of reference images increases the likelihood of including irrelevant entities. Both factors can distract the model’s attention and hinder the generation process. Besides, considering token efficiency, the marginal improvement is negligible: total input tokens increase substantially from 2-Rounds to 3-Rounds, while the retrieval gain remains minimal.

\textbf{Effectiveness of ORIG Components.} Results in Table~\ref{tab:ablation} indicate that removing Knowledge Accumulation and Fine-grained Multimodal Refinement leads to a notable decline in the generation accuracy of Gemini-Image, when using retrieval results from GPT-5 and Qwen2.5-VL-72B as inputs. Specifically, the generation accuracy decreases by 2.8\% and 3.4\% with GPT-based retrieval, and by 2.5\% and 3.0\% with Qwen-based retrieval. These results highlight the importance of the Coarse-to-Fine Filtering process in excluding irrelevant or noisy information before generation. Removing Prompt Extension also degrades performance, with 2.5\% and 2.4\% reductions under GPT- and Qwen-based retrieval, respectively, confirming its role in enriching prompts and aligning retrieved evidence with generation objectives. Furthermore, removing Bootstrapping Retrieval causes a smaller yet consistent drop in performance, with 1.2\% and 0.7\% decreases for GPT- and Qwen-based retrieval, respectively. This suggests that the warm-up retrieval helps stabilize early retrieval, particularly for rare or semantically ambiguous entities.

\begin{table}[t]
    \centering
    \small
    \caption{Accuracy (\%) based on Gemini-Image of different ablation variants. The values after the slash ``/'' denote the performance drop compared to the ORIG.}
    \vspace{-0.3cm}
    \scalebox{0.85}{
    \begin{tabular}{l|cc}
    \toprule
    \textbf{Methods}                & \textbf{GPT-5}  & \textbf{Qwen2.5-VL}\\
    \midrule
    ORIG                            & 51.4          & 41.6 \\
    w.o. Bootstrapping Retrieval                  & 50.2/1.2      & 40.9/0.7\\
    w.o. Knowledge Accumulation           & 47.4/2.8      & 39.1/2.5\\
    w.o. Fine-grained Refinement    & 46.8/3.4      & 38.6/3.0\\
    w.o. Prompt Extension              & 48.9/2.5      & 39.2/2.4\\ 
    \bottomrule
    \end{tabular}}
     \vspace{-0.5cm}
    \label{tab:ablation}
\end{table}

\section{Conclusion}
This work tackles the problem of factual inconsistency in image generation, where outputs often diverge from verifiable knowledge. We formalized Factual Image Generation (FIG) as a new task requiring both visual realism and factual grounding, and proposed ORIG, an open multimodal retrieval-augmented framework that dynamically integrates textual and visual evidence. We further built FIG-Eval, a benchmark for evaluating this task across perceptual, compositional, and temporal dimensions. Experimental results show that ORIG improves factual consistency and generation quality over conventional approaches.
Future work will explore deeper reasoning over retrieved knowledge and finer alignment between semantic intent and visual details.

\bibliography{custom}

\begin{thebibliography}{39}
\providecommand{\natexlab}[1]{#1}

\bibitem[{Bai et~al.(2024)Bai, Wang, Xiao, He, Han, Zhang, and Shou}]{bai2024hallucination}
Zechen Bai, Pichao Wang, Tianjun Xiao, Tong He, Zongbo Han, Zheng Zhang, and Mike~Zheng Shou. 2024.
\newblock Hallucination of multimodal large language models: A survey.
\newblock \emph{arXiv preprint arXiv:2404.18930}.

\bibitem[{Batifol et~al.(2025)Batifol, Blattmann, Boesel, Consul, Diagne, Dockhorn, English, English, Esser, Kulal et~al.}]{batifol2025flux}
Stephen Batifol, Andreas Blattmann, Frederic Boesel, Saksham Consul, Cyril Diagne, Tim Dockhorn, Jack English, Zion English, Patrick Esser, Sumith Kulal, and 1 others. 2025.
\newblock Flux. 1 kontext: Flow matching for in-context image generation and editing in latent space.
\newblock \emph{arXiv e-prints}, pages arXiv--2506.

\bibitem[{Chen et~al.(2021)Chen, Wang, and Wang}]{chen2021dataset}
Wenhu Chen, Xinyi Wang, and William~Yang Wang. 2021.
\newblock A dataset for answering time-sensitive questions.
\newblock \emph{arXiv preprint arXiv:2108.06314}.

\bibitem[{Esser et~al.(2024)Esser, Kulal, Blattmann, Entezari, M{\"u}ller, Saini, Levi, Lorenz, Sauer, Boesel et~al.}]{esser2024scaling}
Patrick Esser, Sumith Kulal, Andreas Blattmann, Rahim Entezari, Jonas M{\"u}ller, Harry Saini, Yam Levi, Dominik Lorenz, Axel Sauer, Frederic Boesel, and 1 others. 2024.
\newblock Scaling rectified flow transformers for high-resolution image synthesis.
\newblock In \emph{Forty-first International Conference on Machine Learning}.

\bibitem[{Fu et~al.(2024)Fu, Hu, Du, Wang, Yang, and Gan}]{fuguiding}
Tsu-Jui Fu, Wenze Hu, Xianzhi Du, William~Yang Wang, Yinfei Yang, and Zhe Gan. 2024.
\newblock Guiding instruction-based image editing via multimodal large language models.
\newblock In \emph{The Twelfth International Conference on Learning Representations}.

\bibitem[{Fu et~al.(2025)Fu, He, Lu, Wang, and Roth}]{fucommonsense}
Xingyu Fu, Muyu He, Yujie Lu, William~Yang Wang, and Dan Roth. 2025.
\newblock Commonsense-t2i challenge: Can text-to-image generation models understand commonsense?
\newblock In \emph{First Conference on Language Modeling}.

\bibitem[{Geng et~al.(2025)Geng, Wang, Ma, Li, Rao, Gu, Zhong, Lu, Hu, Zhang et~al.}]{geng2025x}
Zigang Geng, Yibing Wang, Yeyao Ma, Chen Li, Yongming Rao, Shuyang Gu, Zhao Zhong, Qinglin Lu, Han Hu, Xiaosong Zhang, and 1 others. 2025.
\newblock X-omni: Reinforcement learning makes discrete autoregressive image generative models great again.
\newblock \emph{arXiv preprint arXiv:2507.22058}.

\bibitem[{Hu et~al.(2024{\natexlab{a}})Hu, Chan, Su, Chen, Li, Sohn, Zhao, Ben, Gong, Cohen et~al.}]{hu2024instruct}
Hexiang Hu, Kelvin~CK Chan, Yu-Chuan Su, Wenhu Chen, Yandong Li, Kihyuk Sohn, Yang Zhao, Xue Ben, Boqing Gong, William Cohen, and 1 others. 2024{\natexlab{a}}.
\newblock Instruct-imagen: Image generation with multi-modal instruction.
\newblock In \emph{Proceedings of the IEEE/CVF conference on computer vision and pattern recognition}, pages 4754--4763.

\bibitem[{Hu et~al.(2024{\natexlab{b}})Hu, Gu, Dou, Fayyaz, Lu, Chang, and Peng}]{humrag}
Wenbo Hu, Jia-Chen Gu, Zi-Yi Dou, Mohsen Fayyaz, Pan Lu, Kai-Wei Chang, and Nanyun Peng. 2024{\natexlab{b}}.
\newblock Mrag-bench: Vision-centric evaluation for retrieval-augmented multimodal models.
\newblock In \emph{The Thirteenth International Conference on Learning Representations}.

\bibitem[{Huang et~al.(2025{\natexlab{a}})Huang, Duan, Sun, Xie, Li, and Liu}]{huang2025t2i}
Kaiyi Huang, Chengqi Duan, Kaiyue Sun, Enze Xie, Zhenguo Li, and Xihui Liu. 2025{\natexlab{a}}.
\newblock T2i-compbench++: An enhanced and comprehensive benchmark for compositional text-to-image generation.
\newblock \emph{IEEE Transactions on Pattern Analysis and Machine Intelligence}.

\bibitem[{Huang et~al.(2025{\natexlab{b}})Huang, Yu, Ma, Zhong, Feng, Wang, Chen, Peng, Feng, Qin et~al.}]{huang2025survey}
Lei Huang, Weijiang Yu, Weitao Ma, Weihong Zhong, Zhangyin Feng, Haotian Wang, Qianglong Chen, Weihua Peng, Xiaocheng Feng, Bing Qin, and 1 others. 2025{\natexlab{b}}.
\newblock A survey on hallucination in large language models: Principles, taxonomy, challenges, and open questions.
\newblock \emph{ACM Transactions on Information Systems}, 43(2):1--55.

\bibitem[{Huang et~al.(2024)Huang, He, Long, Wang, Li, Yu, Shu, Chan, Jiang, Wu et~al.}]{huang2024t2i}
Ziwei Huang, Wanggui He, Quanyu Long, Yandi Wang, Haoyuan Li, Zhelun Yu, Fangxun Shu, Long Chan, Hao Jiang, Fei Wu, and 1 others. 2024.
\newblock T2i-factualbench: Benchmarking the factuality of text-to-image models with knowledge-intensive concepts.
\newblock \emph{arXiv preprint arXiv:2412.04300}.

\bibitem[{Kalai and Vempala(2024)}]{kalai2024calibrated}
Adam~Tauman Kalai and Santosh~S Vempala. 2024.
\newblock Calibrated language models must hallucinate.
\newblock In \emph{Proceedings of the 56th Annual ACM Symposium on Theory of Computing}, pages 160--171.

\bibitem[{Karras et~al.(2019)Karras, Laine, and Aila}]{karras2019style}
Tero Karras, Samuli Laine, and Timo Aila. 2019.
\newblock A style-based generator architecture for generative adversarial networks.
\newblock In \emph{Proceedings of the IEEE/CVF conference on computer vision and pattern recognition}, pages 4401--4410.

\bibitem[{Labs(2024)}]{flux2024}
Black~Forest Labs. 2024.
\newblock Flux.
\newblock \url{https://github.com/black-forest-labs/flux}.

\bibitem[{Li et~al.(2024{\natexlab{a}})Li, Lin, Pathak, Li, Fei, Wu, Xia, Zhang, Neubig, and Ramanan}]{li2024evaluating}
Baiqi Li, Zhiqiu Lin, Deepak Pathak, Jiayao Li, Yixin Fei, Kewen Wu, Xide Xia, Pengchuan Zhang, Graham Neubig, and Deva Ramanan. 2024{\natexlab{a}}.
\newblock Evaluating and improving compositional text-to-visual generation.
\newblock In \emph{Proceedings of the IEEE/CVF Conference on Computer Vision and Pattern Recognition}, pages 5290--5301.

\bibitem[{Li et~al.(2024{\natexlab{b}})Li, Xu, Liu, and Xiao}]{li2024unimo}
Wei Li, Xue Xu, Jiachen Liu, and Xinyan Xiao. 2024{\natexlab{b}}.
\newblock Unimo-g: Unified image generation through multimodal conditional diffusion.
\newblock In \emph{Proceedings of the 62nd Annual Meeting of the Association for Computational Linguistics (Volume 1: Long Papers)}, pages 6173--6188.

\bibitem[{Li et~al.(2024{\natexlab{c}})Li, Li, Wang, Jiang, Zhang, Zheng, Wang, Zheng, Huang, Zhou et~al.}]{li2024benchmarking}
Yangning Li, Yinghui Li, Xinyu Wang, Yong Jiang, Zhen Zhang, Xinran Zheng, Hui Wang, Hai-Tao Zheng, Fei Huang, Jingren Zhou, and 1 others. 2024{\natexlab{c}}.
\newblock Benchmarking multimodal retrieval augmented generation with dynamic vqa dataset and self-adaptive planning agent.
\newblock \emph{arXiv preprint arXiv:2411.02937}.

\bibitem[{Liang et~al.(2025)Liang, Xiang, Yu, Zhang, Hong, Fan, and Tang}]{openmanus2025}
Xinbin Liang, Jinyu Xiang, Zhaoyang Yu, Jiayi Zhang, Sirui Hong, Sheng Fan, and Xiao Tang. 2025.
\newblock \href {https://doi.org/10.5281/zenodo.15186407} {Openmanus: An open-source framework for building general ai agents}.

\bibitem[{Lim et~al.(2025)Lim, Choi, and Shim}]{lim2025evaluating}
Youngsun Lim, Hojun Choi, and Hyunjung Shim. 2025.
\newblock Evaluating image hallucination in text-to-image generation with question-answering.
\newblock In \emph{Proceedings of the AAAI Conference on Artificial Intelligence}, volume~39, pages 26290--26298.

\bibitem[{Liu et~al.(2023)Liu, Wu, Zhai, Yuan, and Zhang}]{liu2023riatig}
Han Liu, Yuhao Wu, Shixuan Zhai, Bo~Yuan, and Ning Zhang. 2023.
\newblock Riatig: Reliable and imperceptible adversarial text-to-image generation with natural prompts.
\newblock In \emph{Proceedings of the IEEE/CVF Conference on Computer Vision and Pattern Recognition}, pages 20585--20594.

\bibitem[{Niu et~al.(2025)Niu, Ning, Zheng, Lin, Jin, Liao, Ning, Zhu, and Yuan}]{niu2025wise}
Yuwei Niu, Munan Ning, Mengren Zheng, Bin Lin, Peng Jin, Jiaqi Liao, Kunpeng Ning, Bin Zhu, and Li~Yuan. 2025.
\newblock Wise: A world knowledge-informed semantic evaluation for text-to-image generation.
\newblock \emph{arXiv preprint arXiv:2503.07265}.

\bibitem[{Peng et~al.(2025)Peng, Cui, Tang, Qi, Dong, Bai, Han, Ge, Zhang, and Xia}]{pengdreambench++}
Yuang Peng, Yuxin Cui, Haomiao Tang, Zekun Qi, Runpei Dong, Jing Bai, Chunrui Han, Zheng Ge, Xiangyu Zhang, and Shu-Tao Xia. 2025.
\newblock Dreambench++: A human-aligned benchmark for personalized image generation.
\newblock In \emph{The Thirteenth International Conference on Learning Representations}.

\bibitem[{Qu et~al.(2025)Qu, Li, Wang, Wang, Li, Nie, and Chua}]{qutiger}
Leigang Qu, Haochuan Li, Tan Wang, Wenjie Wang, Yongqi Li, Liqiang Nie, and Tat-Seng Chua. 2025.
\newblock Tiger: Unifying text-to-image generation and retrieval with large multimodal models.
\newblock In \emph{The Thirteenth International Conference on Learning Representations}.

\bibitem[{Ramesh et~al.(2022)Ramesh, Dhariwal, Nichol, Chu, and Chen}]{ramesh2022hierarchical}
Aditya Ramesh, Prafulla Dhariwal, Alex Nichol, Casey Chu, and Mark Chen. 2022.
\newblock Hierarchical text-conditional image generation with clip latents.
\newblock \emph{arXiv preprint arXiv:2204.06125}, 1(2):3.

\bibitem[{Robertson(2024)}]{robertson2024google}
Adi Robertson. 2024.
\newblock Google apologizes for ‘missing the mark’after gemini generated racially diverse nazis.
\newblock \emph{the Verge}, 21.

\bibitem[{Rombach et~al.(2022)Rombach, Blattmann, Lorenz, Esser, and Ommer}]{rombach2022high}
Robin Rombach, Andreas Blattmann, Dominik Lorenz, Patrick Esser, and Bj{\"o}rn Ommer. 2022.
\newblock High-resolution image synthesis with latent diffusion models.
\newblock In \emph{Proceedings of the IEEE/CVF conference on computer vision and pattern recognition}, pages 10684--10695.

\bibitem[{Saharia et~al.(2022)Saharia, Chan, Saxena, Li, Whang, Denton, Ghasemipour, Gontijo~Lopes, Karagol~Ayan, Salimans et~al.}]{saharia2022photorealistic}
Chitwan Saharia, William Chan, Saurabh Saxena, Lala Li, Jay Whang, Emily~L Denton, Kamyar Ghasemipour, Raphael Gontijo~Lopes, Burcu Karagol~Ayan, Tim Salimans, and 1 others. 2022.
\newblock Photorealistic text-to-image diffusion models with deep language understanding.
\newblock \emph{Advances in neural information processing systems}, 35:36479--36494.

\bibitem[{Shalev-Arkushin et~al.(2025)Shalev-Arkushin, Gal, Bermano, and Fried}]{shalev2025imagerag}
Rotem Shalev-Arkushin, Rinon Gal, Amit~H Bermano, and Ohad Fried. 2025.
\newblock Imagerag: Dynamic image retrieval for reference-guided image generation.
\newblock \emph{arXiv preprint arXiv:2502.09411}.

\bibitem[{Wang et~al.(2024)Wang, Zhang, Luo, Sun, Cui, Wang, Zhang, Wang, Li, Yu et~al.}]{wang2024emu3}
Xinlong Wang, Xiaosong Zhang, Zhengxiong Luo, Quan Sun, Yufeng Cui, Jinsheng Wang, Fan Zhang, Yueze Wang, Zhen Li, Qiying Yu, and 1 others. 2024.
\newblock Emu3: Next-token prediction is all you need.
\newblock \emph{arXiv preprint arXiv:2409.18869}.

\bibitem[{Wang et~al.(2025)Wang, Yang, Zhao, Zhang, Liu, Zhou, and Xie}]{wang2025gpt}
Yuhan Wang, Siwei Yang, Bingchen Zhao, Letian Zhang, Qing Liu, Yuyin Zhou, and Cihang Xie. 2025.
\newblock Gpt-image-edit-1.5 m: A million-scale, gpt-generated image dataset.
\newblock \emph{arXiv preprint arXiv:2507.21033}.

\bibitem[{Wu et~al.(2025)Wu, Li, Zhou, Lin, Gao, Yan, Yin, Bai, Xu, Chen et~al.}]{wu2025qwen}
Chenfei Wu, Jiahao Li, Jingren Zhou, Junyang Lin, Kaiyuan Gao, Kun Yan, Sheng-ming Yin, Shuai Bai, Xiao Xu, Yilei Chen, and 1 others. 2025.
\newblock Qwen-image technical report.
\newblock \emph{arXiv preprint arXiv:2508.02324}.

\bibitem[{Xu et~al.(2023)Xu, Liu, Wu, Tong, Li, Ding, Tang, and Dong}]{xu2023imagereward}
Jiazheng Xu, Xiao Liu, Yuchen Wu, Yuxuan Tong, Qinkai Li, Ming Ding, Jie Tang, and Yuxiao Dong. 2023.
\newblock Imagereward: Learning and evaluating human preferences for text-to-image generation.
\newblock \emph{Advances in Neural Information Processing Systems}, 36:15903--15935.

\bibitem[{Xu et~al.(2025)Xu, Chen, Lyu, Zhao, Xiong, Lin, Han, and Ding}]{xu2025mitigating}
Xinhao Xu, Hui Chen, Mengyao Lyu, Sicheng Zhao, Yizhe Xiong, Zijia Lin, Jungong Han, and Guiguang Ding. 2025.
\newblock Mitigating hallucinations in multi-modal large language models via image token attention-guided decoding.
\newblock In \emph{Proceedings of the 2025 Conference of the Nations of the Americas Chapter of the Association for Computational Linguistics: Human Language Technologies (Volume 1: Long Papers)}, pages 1571--1590.

\bibitem[{Yan et~al.(2025)Yan, Ye, Li, Huang, Yuan, He, Lin, He, He, and Yuan}]{yan2025gpt}
Zhiyuan Yan, Junyan Ye, Weijia Li, Zilong Huang, Shenghai Yuan, Xiangyang He, Kaiqing Lin, Jun He, Conghui He, and Li~Yuan. 2025.
\newblock Gpt-imgeval: A comprehensive benchmark for diagnosing gpt4o in image generation.
\newblock \emph{arXiv preprint arXiv:2504.02782}.

\bibitem[{Yuan et~al.(2025)Yuan, Zhao, Wang, Xiao, Ni, Liu, and Dou}]{yuan2025finerag}
Huaying Yuan, Ziliang Zhao, Shuting Wang, Shitao Xiao, Minheng Ni, Zheng Liu, and Zhicheng Dou. 2025.
\newblock Finerag: Fine-grained retrieval-augmented text-to-image generation.
\newblock In \emph{Proceedings of the 31st International Conference on Computational Linguistics}, pages 11196--11205.

\bibitem[{Zhang et~al.(2017)Zhang, Xu, Li, Zhang, Wang, Huang, and Metaxas}]{zhang2017stackgan}
Han Zhang, Tao Xu, Hongsheng Li, Shaoting Zhang, Xiaogang Wang, Xiaolei Huang, and Dimitris~N Metaxas. 2017.
\newblock Stackgan: Text to photo-realistic image synthesis with stacked generative adversarial networks.
\newblock In \emph{Proceedings of the IEEE international conference on computer vision}, pages 5907--5915.

\bibitem[{Zhao et~al.(2024)Zhao, Song, Wang, Feng, Ding, Sun, Xiao, and Wang}]{zhao2024monoformer}
Chuyang Zhao, Yuxing Song, Wenhao Wang, Haocheng Feng, Errui Ding, Yifan Sun, Xinyan Xiao, and Jingdong Wang. 2024.
\newblock Monoformer: One transformer for both diffusion and autoregression.
\newblock \emph{arXiv preprint arXiv:2409.16280}.

\bibitem[{Zhou et~al.(2024)Zhou, YU, Babu, Tirumala, Yasunaga, Shamis, Kahn, Ma, Zettlemoyer, and Levy}]{zhoutransfusion}
Chunting Zhou, LILI YU, Arun Babu, Kushal Tirumala, Michihiro Yasunaga, Leonid Shamis, Jacob Kahn, Xuezhe Ma, Luke Zettlemoyer, and Omer Levy. 2024.
\newblock Transfusion: Predict the next token and diffuse images with one multi-modal model.
\newblock In \emph{The Thirteenth International Conference on Learning Representations}.

\end{thebibliography}

\clearpage
\appendix

\section{More Implementation Details}

\begin{table}[t]
\centering
\small
\caption{Cost and time analysis across pipeline stages.}
\vspace{-0.3cm}
\scalebox{0.65}{
\begin{tabular}{l|rrrrr}
\toprule
\textbf{Metric} &
\makecell[c]{\textbf{Boots-}\\\textbf{trapping}\\\textbf{retrieval}} &
\makecell[c]{\textbf{Query}\\\textbf{Planning}} &
\makecell[c]{\textbf{Knowledge}\\\textbf{Accumu-}\\\textbf{lation}} &
\makecell[c]{\textbf{Fine-}\\\textbf{grained}\\\textbf{Refine}} &
\makecell[c]{\textbf{Prompt}\\\textbf{Extension}} \\
\midrule
Total Retrieval & 1.2 & 6.7 & -- & -- & -- \\
Retrieval Time (s) & 1.1 & 7.7 & -- & -- & -- \\
Input Tokens & 61.6 & 1{,}021.1 & 3{,}598.8 & 827.1 & 740.6 \\
Output Tokens & 102.6 & 138.8 & 827.1 & 740.6 & 519.1 \\
\bottomrule
\end{tabular}}
\label{tab:cost_time}
\vspace{-0.5cm}
\end{table}

During retrieval, textual and visual queries were processed separately to ensure modality-specific optimization. For text retrieval, we used the Serper API to access Google Search (region = US, language = English), retrieving the top 10 webpage results per query, each containing a URL and a short textual snippet. To prevent redundant crawling and enhance relevance, the snippets were evaluated by the retrieval backbone model based on semantic similarity, and the top 2 most relevant webpages were retained. The full content of these pages was then extracted using the Jina.ai Reader API, providing clean textual evidence for subsequent reasoning and grounding. For image retrieval, we used Serper’s Google Image Search endpoint to obtain the top 10 image results per query. We selected the top 5 accessible and unique images after removing broken or duplicate links, ensuring both coverage and diversity of visual evidence. The retrieval backbone was Qwen2.5-VL, applied in multi-scale variants (72B, 32B, and 7B), complemented by a GPT-5 (2025.10) retriever with default sampling parameters (temperature $\le$ 0.2, top-p default). For image generation, we employed Gemini-Image (Gemini-2.5-Flash-Image, “Nano-Banana” version), GPT-Image-1, and Qwen-Image, all accessed through their official APIs with default configurations consistent with their released repositories, ensuring comparability across generation backbones. All instructions, prompts, evaluation scripts, and annotated references will be released in the anonymous GitHub repository.

\subsection{Token Usage}
Table \ref{tab:cost_time} reports the cost and time statistics across different pipeline stages. Average Search denotes the total number of text and image retrieval calls, while Input Tokens correspond to the average number of tokens fed into the retrieval backbone at each stage. The Query Planning phase shows relatively high input token usage because each planning step must condition on previously accumulated information within the knowledge database to decide the next retrieval direction. The Knowledge Accumulation stage consumes the largest number of tokens, as it involves processing and filtering the full multimodal content of the retrieved information. 

Although the pipeline is token-intensive, this higher cost is intrinsic to the FIG task itself, which requires inherently heavy retrieval for multimodal, knowledge-grounded image generation. Such generation must render fine-grained scene details using complementary textual and visual evidence, thereby necessitating richer and more exhaustive retrieval than typical VQA-style tasks that seek only a single textual answer. To maintain practicality, several safeguards are integrated into the pipeline. Adaptive query generation dynamically produces one to five modality-specific retrieval queries per round, balancing coverage and efficiency (Table \ref{tab:retrieval_cost}). Sufficiency-based iteration control halts retrieval once adequate evidence has been gathered, preventing redundant web expansion and mitigating downstream token growth.

\begin{table}[t]
\centering
\small
\caption{Retrieval cost with different iterations. Text retrieval represents the total retrieval times of Google Search, where Image Retrieval represents the retrieval times of Google Image, Input and Output Tokens is the length of input and output for prompt extension stage.}
\vspace{-0.3cm}
\scalebox{0.8}{
\begin{tabular}{lccrr}
\toprule
\textbf{Iter.} &
\makecell[c]{\textbf{Text}\\\textbf{Retrieval}} &
\makecell[c]{\textbf{Image}\\\textbf{Retrieval}} &
\makecell[c]{\textbf{Input}\\\textbf{Tokens}} &
\makecell[c]{\textbf{Output}\\\textbf{Tokens}} \\
\midrule
+1-Round & 2.9 & 1.8 & 657.9 & 453.5 \\
+2-Round & 4.6 & 2.7 & 784.8 & 536.4 \\
+3-Round & 6.3 & 3.2 & 941.1 & 597.2 \\
ORIG     & 4.5 & 2.5 & 740.6 & 527.1 \\
\bottomrule
\end{tabular}}
\label{tab:retrieval_cost}
\vspace{-0.5cm}
\end{table}

\begin{table*}[t]
\centering
\caption{Accuracy(\%) comparison based on Qwen2.5-VL-72B retrieved evidence across 10 entity classes and 3 concept categories on the FIG-Eval dataset. All represents the average score of all 10 categories.}
\label{tab:ablation1}
\vspace{-0.3cm}
\small
\scalebox{0.92}{
\begin{tabular}{l|cccccccccc|ccc|c}
\toprule
\textbf{Methods} & \textbf{An.} & \textbf{Sp.} & \textbf{Tr.} & \textbf{La.} & \textbf{Fo.} & \textbf{Pe.} & \textbf{Pl.} & \textbf{Pr.} & \textbf{Cu.} & \textbf{Ev.} & \textbf{PF} & \textbf{CC} & \textbf{TC} & \textbf{All} \\
\midrule
Qwen-Image                    & 16.4 & 9.3  & 19.3 & 31.3 & 12.8 & 16.7 & 23.3 & 12.3 & 30.2 & 17.7 & 21.5 & 18.2 & 14.0 & 19.0 \\
\hspace{0.5em}Prompt Enhanced & 33.5&	12.7&	25.3&	39.2&	19.7&	21.9&	37.8&	24.0&	37.6&	27.3&	28.4&	27.2&	30.4&	28.1\\
\hspace{0.5em}ORIG            &36.6&	21.1&	41.5&	45.1&	28.2&	23.6&	50.1&	37.7&	42.0&	33.9&	37.1&	36.5&	33.6&	36.1\\
\hspace{0.5em}ORIG-Img        &33.8&	16.7&	37.2&	39.8&	20.1&	22.8&	47.3&	33.5&	35.7&	31.0&	34.8&	31.0&	27.8&	32.1\\
\hspace{0.5em}ORIG-Txt        &35.6&	18.7&	35.3&	40.6&	24.1&	23.5&	48.5&	28.7&	40.8&	31.2&	33.3&	34.0&	30.9&	32.9\\
\midrule
Gemini-Image                  & 47.0 & 20.0 & 32.3 & 45.2 & 31.0 & 22.3 & 51.3 & 22.6 & 42.5 & 30.4 & 34.9 & 34.4 & 35.1 & 34.6\\
\hspace{0.5em}Prompt Enhanced &38.4&	24.0&	33.8&	44.0&	22.9&	28.0&	51.6&	28.3&	41.3&	34.9&	35.4&	33.8&	37.2&	34.9 \\
\hspace{0.5em}ORIG            &47.9&	26.4&	43.9&	50.8&	29.6&	33.3&	53.3&	42.7&	41.9&	44.4&	42.5&	41.8&	40.0&	41.6\\
\hspace{0.5em}ORIG-Img        &46.2&	21.3&	37.5&	44.7&	25.2&	27.0&	51.0&	37.2&	41.4&	37.4&	39.6&	35.6&	35.9&	37.3\\
\hspace{0.5em}ORIG-Txt        &48.0&	24.6&	38.1&	45.6&	23.9&	28.2&	52.4&	32.4&	41.3&	40.3&	37.3&	38.6&	38.8&	37.8\\
\midrule
 GPT-Image                    & 39.7 & 17.1 & 33.1 & 44.5 & 21.8 & 20.2 & 45.7 & 31.9 & 35.1 & 30.3 & 34.6 & 30.7 & 29.1 & 32.1\\
 \hspace{0.5em}Prompt Enhanced&35.7&	   21.9&	31.1&	41.7&	18.6&	26.7&	49.1&	25.7&	39.6&	33.0&	33.7&	31.8&	31.8&	32.5 \\
\hspace{0.5em}ORIG            &44.6&	25.1&	40.2&	49.8&	26.1&	33.8&	51.6&	45.3&	41.4&	44.4&	42.6&	38.4&	40.5&	40.5\\
\hspace{0.5em}ORIG-Img        &42.7&	22.1&	38.4&	42.6&	23.9&	30.8&	47.9&	33.9&	39.1&	35.8&	38.7&	33.8&	35.1&	36.1\\
\hspace{0.5em}ORIG-Txt        &44.3&	23.9&	35.9&	45.2&	24.3&	30.1&	49.9&	34.8&	40.5&	39.2&	37.4&	36.7&	38.9&	37.2\\
\bottomrule
\end{tabular}}
\vspace{-0.2cm}
\end{table*}

\begin{table*}[t]
\centering
\caption{Accuracy(\%) comparison with Qwen2.5-VL-72B/-32B/-7B retrieved evidence across 10 entity classes and 3 concept categories on the FIG-Eval dataset. All represents the average score of all 10 categories. The image generation model is Gemini-Image}
\label{tab:ablation2}
\vspace{-0.3cm}
\small
\scalebox{0.92}{
\begin{tabular}{l|cccccccccc|ccc|c}
\toprule
\textbf{Methods} & \textbf{An.} & \textbf{Sp.} & \textbf{Tr.} & \textbf{La.} & \textbf{Fo.} & \textbf{Pe.} & \textbf{Pl.} & \textbf{Pr.} & \textbf{Cu.} & \textbf{Ev.} & \textbf{PF} & \textbf{CC} & \textbf{TC} & \textbf{All} \\
\midrule
\multicolumn{15}{c}{Qwen2.5-VL-72B} \\
\midrule
Prompt Enhanced &38.4&	24.0&	33.8&	44.0&	22.9&	28.0&	51.6&	28.3&	41.3&	34.9&	35.4&	33.8&	37.2&	34.9 \\
ORIG            &47.9&	26.4&	43.9&	50.8&	29.6&	33.3&	53.3&	42.7&	41.9&	44.4&	42.5&	41.8&	40.0&	41.6\\
\midrule
\multicolumn{15}{c}{Qwen2.5-VL-32B} \\
\midrule
Prompt Enhanced & 38.5&	24.3&	33.5&	42.4&	22.9&	27.6&	51.9&	27.7&	38.1&	34.6&	34.6&	34.6&	34.8&	34.4\\
ORIG            &43.1&	26.1&	37.4&	47.7&	28.4&	33.7&	53.0&	38.0&	41.3&	42.3&	40.6&	39.5&	37.3&	39.5\\
\midrule
\multicolumn{15}{c}{Qwen2.5-VL-7B} \\
\midrule
Prompt Enhanced & 38.2&	22.7&	33.0&	40.2&	22.7&	26.7&	50.4&	26.3&	36.5&	33.8&	34.4&	32.6&	33.1&	33.4\\
ORIG            &40.8&	23.6&	35.1&	45.5&	26.3&	31.9&	52.0&	36.6&	38.9&	37.4&	38.0&	37.2&	35.9&	37.2\\
\bottomrule
\end{tabular}}
\vspace{-0.4cm}
\end{table*}

\section{Supplement to Experiments}
\subsection{Results Based on Qwen-Retrieval}

As shown in Table~\ref{tab:ablation1}, we present detailed results on the FIG-Eval dataset using the Qwen2.5-VL-72B retrieval backbone, covering ten entity classes and three concept categories. The overall trend aligns well with the main results in Table~\ref{tab:main_results}: all three image generation models (Qwen-Image, Gemini-Image, and GPT-Image) exhibit substantial accuracy improvements after integrating retrieved external evidence, while the full \textbf{ORIG} framework consistently achieves the highest scores across nearly all categories. The most pronounced gains are observed in semantically complex and frequently updated domains such as Products, ransportation, and Events, highlighting the crucial role of factual and time-sensitive web evidence in enhancing the fidelity and contextual correctness of image generation. Furthermore, improvements in Temporal Consistency and Compositional Consistency demonstrate that retrieval-grounded generation better preserves coherent temporal and relational structures that purely parametric generation often fails to capture. A comparison between the \textbf{ORIG-Img} and \textbf{ORIG-Txt} variants further verifies the importance of multimodal fusion. Text-only retrieval improves context consistency but lacks perceptual grounding, whereas image-only retrieval contributes to visual alignment but fails to capture factual attributes. Their combination within the ORIG framework consistently outperforms both unimodal variants, demonstrating that Qwen2.5-VL-72B effectively fuses complementary textual and visual knowledge. 

\begin{table*}[t]
\centering
\caption{Accuracy comparison (\%) of four models on direct generation task across 10 entity classes and 3 concept categories using the FIG-Eval benchmark. All indicates the average performance across all entity categories.}
\label{tab:ablation3}
\vspace{-0.3cm}
\small
\scalebox{0.92}{
\begin{tabular}{l|cccccccccc|ccc|c}
\toprule
\textbf{Methods} & \textbf{An.} & \textbf{Sp.} & \textbf{Tr.} & \textbf{La.} & \textbf{Fo.} & \textbf{Pe.} & \textbf{Pl.} & \textbf{Pr.} & \textbf{Cu.} & \textbf{Ev.} & \textbf{PF} & \textbf{CC} & \textbf{TC} & \textbf{All} \\
\midrule
Flux-Schnell                    & 13.4 & 4.0  & 18.7 & 22.1 & 6.1 & 12.6 & 9.5 & 14.9 & 13.5 & 24.6 & 14.6 & 13.2 & 12.7 & 13.8 \\
Flux-Dev                    & 9.6 & 10.5  & 12.9 & 27.1 & 5.0 & 15.0 & 12.5 & 20.4 & 16.7 & 21.6 & 18.2 & 13.1 & 9.8 & 14.9 \\
SD-3.5-Large    & 19.3 & 8.4  & 16.3 & 31.1 & 10.7 & 9.9 & 17.3 & 10.5 & 11.5 & 20.4 & 16.8 & 17.5 & 14.8 & 16.8  \\
\bottomrule
\end{tabular}}
\vspace{-0.4cm}
\end{table*}

\subsection{Results Across Different Model Scales}
As shown in Table~\ref{tab:ablation2}, the retrieval performance of the Qwen2.5-VL family scales consistently with model size. Larger variants exhibit stronger reasoning and evidence integration abilities, leading to clear accuracy gains across both entity and concept dimensions. Specifically, Qwen2.5-VL-72B achieves the highest overall accuracy (41.6\%), surpassing the 32B and 7B models by 2.1\% and 4.4\%, respectively. The improvement mainly stems from enhanced cross-modal grounding and better utilization of retrieved knowledge, particularly in semantically complex and frequently updated domains classes such as Products and Events.

\subsection{Additional Generation Results}

As shown in Table~\ref{tab:ablation3}, we additionally evaluate several representative diffusion-based models on the FIG-Eval benchmark. The diffusion-based models such as Flux-Schnell~\citep{flux2024}, Flux-Dev~\citep{flux2024}, and SD-3.5-Large~\citep{esser2024scaling} demonstrate limited performance in factual grounding and multi-entity reasoning. Their accuracy across concept categories remains lower than LLM-based models like Gemini-Image or GPT-Image, indicating that diffusion-based models possess limited intrinsic knowledge and lack the reasoning capabilities required to interpret knowledge-intensive or context-dependent visual semantics.

\subsection{Evaluation of I-HallA Benchmark}
I-HallAbench \cite{lim2025evaluating} (I-HallA) is a benchmark for evaluating image hallucination in text-to-image models, using QA pairs generated by GPT-4o to assess factual accuracy. It includes 1,000 multiple-choice sets to measure how well models reflect factual information in generated images. We tested the performance of ORIG framework on the I-HallA benchmark using Gemini-Image, GPT-Image and Qwen-Image as the image generation models. Although I-HallA focuses on factual domains where strong models such as GPT and Gemini have already internalized most knowledge, thereby reducing the marginal benefit of retrieval (compared to FIG-Eval), our approach still yields consistent and noticeable improvements in generation accuracy (Table \ref{tab:i_halle}), demonstrating strong generalizability beyond our own dataset.

\begin{table}[t]
\centering
\small
\caption{Results (Accuracy \%) on I-HallA, evaluation using I-HallA score. "w. ORIG" represents the ORIG augmented generation process.}
\vspace{-0.3cm}
\scalebox{0.67}{
\begin{tabular}{l|rrrrrr}
\toprule
\textbf{Metric} &
\makecell[c]{\textbf{Gemini-}\\\textbf{Image}} &
\makecell[c]{\textbf{Gemini-}\\\textbf{Image}\\\textbf{w. ORIG}} &
\makecell[c]{\textbf{GPT-}\\\textbf{Image}} &
\makecell[c]{\textbf{GPT-}\\\textbf{Image}\\\textbf{w. ORIG}} &
\makecell[c]{\textbf{Qwen-}\\\textbf{Image}} &
\makecell[c]{\textbf{Qwen-}\\\textbf{Image}\\\textbf{w. ORIG}}\\
\midrule
Science & 0.826 & 0.854 & 0.818 & 0.852 & 0.728 & 0.788 \\
History & 0.806 & 0.834 & 0.796 & 0.822 & 0.667 & 0.712 \\
\bottomrule
\end{tabular}}
\label{tab:i_halle}
\vspace{-0.5cm}
\end{table}

\section{Supplement to FIG-Eval}
\subsection{Comparison with Prior Benchmarks}
As shown in Tables~\ref{tab:mlig_prompt} and~\ref{tab:mlig_retrieval_0}, FIG-Eval differs fundamentally from existing text-to-image benchmarks in its task definition, knowledge dependency, and retrieval formulation. Existing benchmarks primarily assess models’ ability to generate from internal (parametric) knowledge, relying on commonsense priors or style imitation. Reasoning, when required, typically operates within the model’s internal representation without external grounding. In contrast, FIG-Eval explicitly targets open-domain, multimodal knowledge integration. It evaluates whether models can retrieve and compose factual visual and textual evidence to address knowledge-intensive prompts that extend beyond their parametric memory.

Table~\ref{tab:mlig_prompt} compares representative prompts across benchmarks. While datasets such as Instruct-Imagen, UNIMO-G, and DREAMBENCH++ are solvable through internal stylistic or compositional reasoning, FIG-Eval requires external factual grounding. For example, the prompt \textit{“Generate an image of a Tesla robot and a Unitree-G1 robot shaking hands”} cannot be resolved using parametric priors alone, which demands retrieval of textual descriptions (e.g., model specifications, relative heights) and visual appearances from external sources. This design encourages retrieval-dependent reasoning, ensuring that successful generation reflects both linguistic–visual alignment and factual consistency.

Furthermore, as shown in Table~\ref{tab:mlig_retrieval_0}, FIG-Eval adopts a more realistic retrieval setting compared to prior benchmarks. Earlier methods such as ImageRAG, TIGER, and RE-IMAGEN rely solely on image-based retrieval from static, locally stored datasets using CLIP embeddings. By contrast, our ORIG pipeline performs iterative, open-domain retrieval on the web (via Google), jointly leveraging textual and visual evidence with dynamic filtering and sufficiency evaluation. This multimodal retrieval setting not only expands the accessible knowledge space but also provides more faithful, up-to-date evidence for complex real-world prompts. Overall, FIG-Eval bridges the gap between conventional parametric generation and retrieval-augmented multimodal reasoning, providing a comprehensive benchmark for assessing factuality, compositionality, and knowledge grounding in text-to-image models.

\begin{table*}[t]
\centering
\caption{Representative prompt comparison between FIG-Eval and existing benchmarks. Unlike prior datasets solvable through parametric knowledge alone, FIG-Eval requires multimodal evidence-grounded reasoning.}
\label{tab:mlig_prompt}
\vspace{-0.3cm}
\small
\scalebox{0.85}{
\begin{tabular}{l|p{12cm}|c}
\toprule
\textbf{Dataset} & \textbf{Prompt Example} & \textbf{Knowledge Source} \\
\midrule
Instruct-Imagen & Render an image of the [ref\#1] vase that depicts the caption, adopting the style of [ref\#2] style image: a vase with flowers on top. & Internal \\
\midrule
UNIMO-G & \textless image\#1\textgreater\ wearing \textless image\#2\textgreater & Internal \\
\midrule
DREAMBENCH++ & A tiny kitten navigating a vast, otherworldly landscape on a leaf sailboat. & Internal \\
\midrule
RISEBench & Draw what they [ref image] will look like after being kept in a daily environment for a year. & Internal, Reasoning \\
\midrule
Commonsense-T2I & A lightbulb without electricity. & Internal, Reasoning \\
\midrule
\textbf{FIG-Eval} & Generate an image of a Tesla robot and a Unitree-G1 robot shaking hands. & External, Reasoning \\
\bottomrule
\end{tabular}}
\vspace{-0.1cm}
\end{table*}

\begin{table*}[t]
\centering
\caption{Comparison of retrieval settings between FIG-Eval and prior benchmarks.}
\label{tab:mlig_retrieval_0}
\vspace{-0.3cm}
\small
\scalebox{0.9}{
\begin{tabular}{l|p{7cm}|c|c|c}
\toprule
\textbf{Method} & \textbf{Prompt Example} & \textbf{Modalities} & \textbf{Retrieval} & \textbf{Source} \\
\midrule
ImageRAG & Grizzly bear in calculus class. & Image & CLIP & Local dataset \\
TIGER & Mary Poppins flying with balloons. & Image & CLIP & Local dataset \\
RE-IMAGEN & Braque Saint-Germain taking a shower. & Image & CLIP & Local dataset \\
FineRAG & Cat with Santa mug. & Image & CLIP & Local dataset \\
\textbf{ORIG (Ours)} & Tesla and UniTree robots shaking hands. & Image + Text & Google-based & Open Web \\
\bottomrule
\end{tabular}}
\vspace{-0.1cm}
\end{table*}

\subsection{Example of Retrieval Principles} Multi-hop sequential retrieval refers to a staged retrieval process in which the model first performs text retrieval to obtain factual and symbolic information, followed by image retrieval to supplement perceptual details. For example, when generating “the top-3 best-selling electric vehicles in North America this year”, the model first retrieves textual data to identify the specific models and brands (e.g., Tesla Model Y, Ford Mustang Mach-E), and then retrieves corresponding images to capture their visual appearance and structural features, achieving stepwise text-to-image grounding. In contrast, parallel co-retrieval conducts text and image retrieval simultaneously within the same iteration, integrating complementary multimodal evidence for knowledge alignment. For instance, when generating “a comparison image of Nintendo Switch and Switch 2”, the model jointly retrieves textual information (e.g., size differences and hardware specifications) and visual information (device appearance images), enabling the generated output to accurately reflect real-world proportions and design variations.

\subsection{Reference Content and Evaluation}
In this section, we provide examples illustrating the structure of reference content in our dataset, including both textual and visual entries, as well as the evaluation questions derived from them. Each prompt is associated with manually curated reference text and images that reflect the intended semantics and visual context, as shown in Figure~\ref{fig:prompt_length_2}. Based on these references, we construct a set of fine-grained, true/false evaluation questions targeting specific visual attributes or factual elements. This structured annotation process enables consistent and interpretable assessment of generation quality across models and concept categories.

\begin{figure}
\centering
\includegraphics[width=0.98\linewidth]{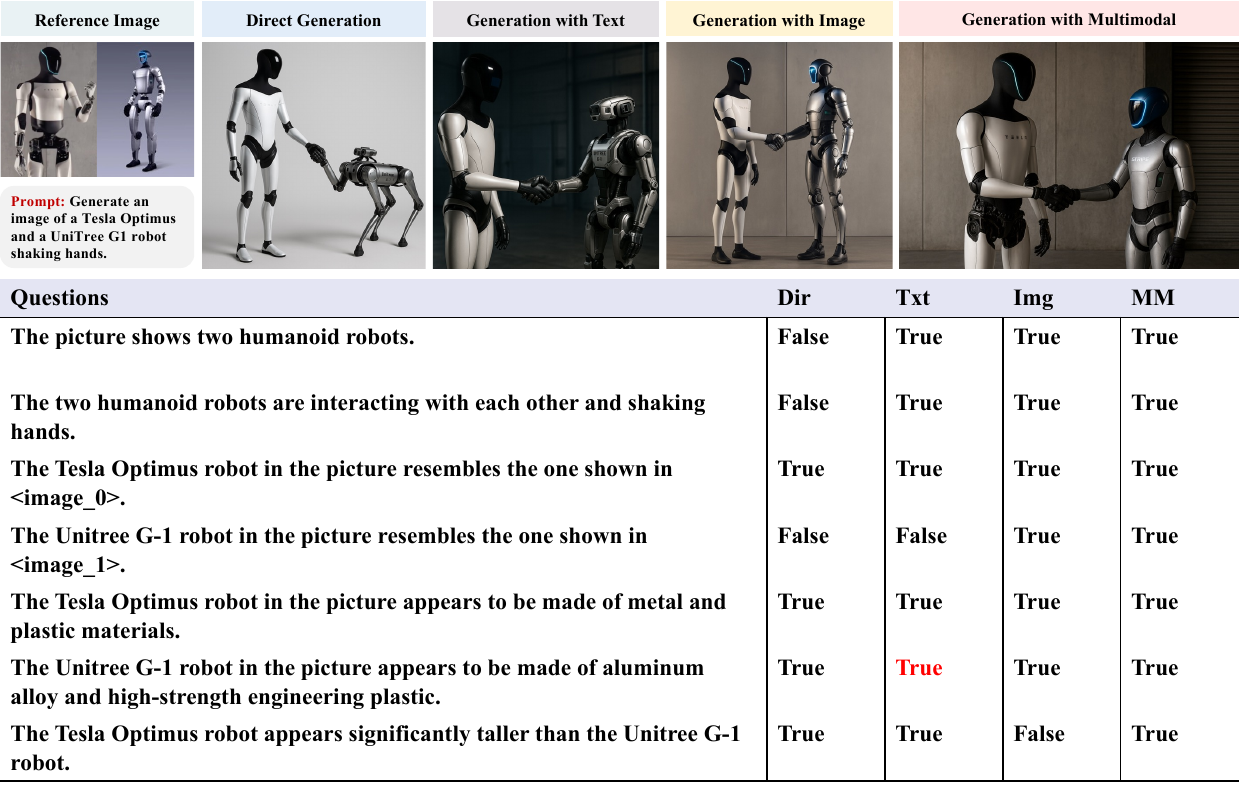}
\caption{Question-answering scores for different modality retrieval tasks (evaluated by GPT-5), with red indicating incorrect evaluations.}
\label{fig:prompt_length_2}
\vspace{-0.5 cm}
\end{figure}

\begin{small}
\begin{tcolorbox}[promptbox, title= Reference for Prompt "Generate an image of a Tesla Optimus and a Unitree-G1 robot
shaking hands."]

\textbf{Reference Text:}
\begin{lstlisting}
1. Two robots: Tesla humanoid robot (Optimus) and Unitree G-1 robot.

2. Optimus and Unitree G-1 both have humanoid structures: human-like form including two arms, two legs, and a torso.

3. Unitree G-1:
- Unitree G-1 robot is approximately 127 cm tall and weighs around 35 kg.
- It is made from aluminum alloy and high-strength engineering plastic.
- The head is a hollow structure with a light ring.

4. Tesla Optimus:
- Tesla Optimus is about 173 cm and weighs 73 kg. 
- constructed using metal and plastic materials.
- Optimus has a display screen on the face for interaction and communication.
\end{lstlisting}

\textbf{Reference Image}
\begin{lstlisting}
1. Img_0 (Tesla Optimus Picture)
- Title: .../- Url: .../- Local Path: ... 
2. Img_1 (Unitree G-1 Picture) ...
\end{lstlisting}

\end{tcolorbox}    
\end{small}

\section{Case Study}

\subsection{Comparison between Different Retrieval Methods}
In addition to the quantitative evaluation presented in Section~\ref{sec:ab_study}, we further compare our method with two representative agentic multimodal retrieval systems (OmniSearch, OpenManus) through case analysis. While Table~\ref{tab:retrieval_methods} reports overall retrieval accuracy across the dataset, here we focus on illustrating the fundamental differences in retrieval objective, evidence usage, and resulting generation quality.

\begin{figure}
\centering
\includegraphics[width=1\linewidth]{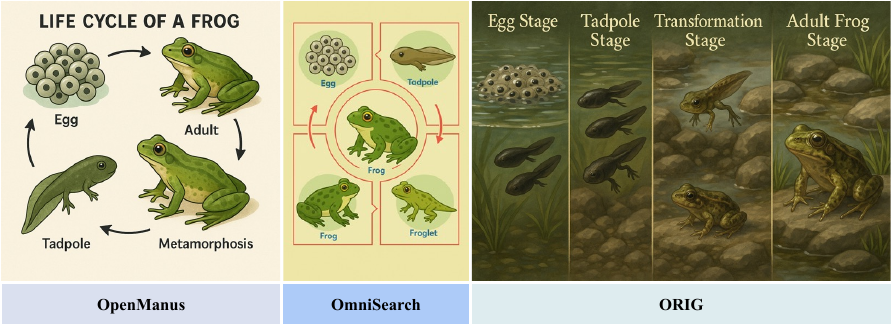}
\caption{The generation results for prompt "Generate a picture of a frog's life cycle" with three retrieval methods}
\label{fig:re_compare}
\vspace{-0.5 cm}
\end{figure}

As shown in Figure~\ref{fig:re_compare}, for the prompt \textit{“Generate a picture of a frog’s life cycle”}, all three methods retrieve relevant external knowledge for expanding the prompt; however, only our approach successfully identifies fine-grained and visually aligned evidence that directly supports accurate image generation. In contrast, OmniSearch and OpenManus often retrieve semantically correct but visually underspecified content, resulting in images that omit critical visual cues, such as developmental stages, spatial configuration, or appearance transitions. This limitation arises because both systems were originally designed for multimodal question answering, where the retrieval target emphasizes semantic sufficiency rather than visual faithfulness. To further clarify the distinction, Table~\ref{tab:mlig_retrieval} provides a structured comparison among the three methods, highlighting their task focus, pipeline design, and knowledge granularity in the FIG-Eval setting. While OpenManus and OmniSearch also employ multimodal retrieval, ORIG fundamentally differs in its task formulation, evidence usage, and pipeline design, yielding more fine-grained and visually grounded outputs.

\textbf{Task Difference.} OmniSearch is optimized for producing a single textual answer, with images serving only as auxiliary or text-converted cues. As a result, its retrieved evidence remains coarse-grained and weakly linked to concrete visual attributes. Similarly, OpenManus focuses on explanatory reasoning, retrieving verbose yet abstract content that seldom translates into explicit visual constraints, rendering much of its retrieved prose redundant for image synthesis. In contrast, ORIG is specifically designed for knowledge-grounded text-to-image generation, where factual and perceptual evidence must be jointly mapped to visual structures. Both textual and visual retrieval results are complementary and jointly essential for producing controllable, faithful images.

\textbf{Pipeline Difference.} ORIG introduces several design innovations absent in prior systems: ORIG introduces several design innovations absent in prior systems, which collectively enable finer-grained retrieval alignment and more coherent generation. First, it begins with bootstrapping retrieval for stable planning, performing an initial lightweight evidence search that helps stabilize sub-query decomposition under rare or ambiguous concepts.
Next, it performs modality-aware knowledge accumulation and filtering, where textual evidence is dynamically refined based on the evolving knowledge state, and image retrieval is subsequently conditioned on the refined textual context. This step preserves complementary information across modalities and enhances cross-modal consistency. Finally, ORIG adopts a two-stage prompt construction process that integrates multimodal content refinement, through summarization and deduplication, along with visual–textual alignment, resulting in concise, generation-ready prompts. This structured pipeline avoids the redundancy and misalignment typical of raw retrieval concatenation, achieving both higher factual accuracy and improved visual coherence.

\begin{table*}[t]
\centering
\caption{Comparison of retrieval settings between FIG-Eval and prior benchmarks.}
\label{tab:mlig_retrieval}
\vspace{-0.3cm}
\small
\scalebox{0.8}{
\begin{tabular}{l|c|c|c}
\toprule
\textbf{Method} & \textbf{Task} & \textbf{Components} & \textbf{Knowledge Granularity }  \\
\midrule
OmniSearch & Knowledge-intensive VQA & Planning; Adaptive retrieval & Coarse \\
OpenManus & Knowledge-intensive VQA & Planning; Adaptive retrieval & Coarse \\
ORIG & FIG & Planning; Adaptive retrieval; Content filtering; Prompt Expanding & Fine-grained \\
\bottomrule
\end{tabular}}
\vspace{-0.3cm}
\end{table*}

\subsection{Effect of Retrieved Text length and details}
This section investigates how the length and granularity of retrieved textual information affect image generation quality. Specifically, we aim to examine whether generation models are capable of reflecting fine-grained details present in the input and whether increasing the level of description improves the output. To explore this, we present two case studies. For each case, we use GPT-5 to extract key content from the retrieved text and construct three image generation prompts at different granularity levels: \textbf{coarse}, \textbf{normal}, and \textbf{fine}. We then compare the generated results across these levels to evaluate the model's ability to incorporate varying degrees of detail.

\begin{figure}
\centering
\includegraphics[width=1\linewidth]{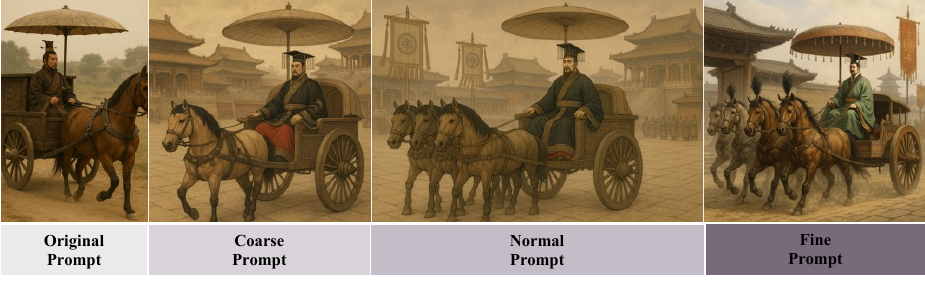}
\caption{Different granularity for prompt "Generate a picture of a Qin Dynasty emperor riding in a horse-drawn carriage."}
\label{fig:prompt_length_1}
\vspace{-0.2 cm}
\end{figure}

We employ GPT-Image as our generation model for all prompt variants. As shown in Figure~\ref{fig:prompt_length_1}, in most cases, increasing the granularity of textual descriptions leads to more accurate and visually aligned generations. The model is generally able to reflect fine-grained visual features, when these details are explicitly included in the prompt. However, we also observe notable failure cases. As illustrated in Figure~\ref{fig:prompt_length_2}, despite progressively refining the description to emphasize that the beetle emits a chemical spray from its rear as a defense mechanism, the model consistently fails to reproduce this behavior accurately in the generated images. Even when we provide an additional ground-truth reference image alongside the prompt, the output remains inconsistent with the described action. This suggests that beyond a certain level of detail, the model’s ability to incorporate fine-grained semantics is limited not by the prompt itself, but by inherent limitations in the model’s capacity for fine-grained visual grounding and compositional fidelity.

\begin{figure}
\centering
\includegraphics[width=1\linewidth]{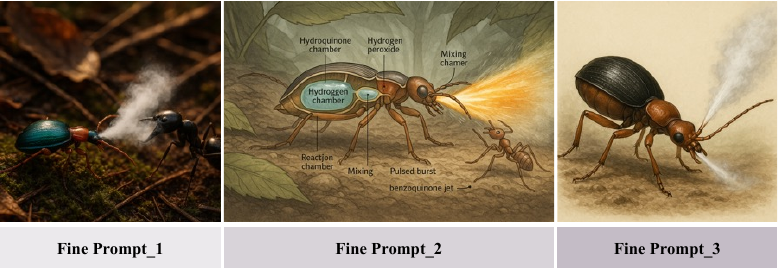}
\caption{Different granularity for prompt "Generate a photo of a bombardier beetle attacking an enemy."}
\label{fig:prompt_length_2}
\vspace{-0.5 cm}
\end{figure}

\subsection{Effect of Retrieved Images}
\begin{figure*}
\centering
\scalebox{0.9}{
\includegraphics[width=1\textwidth]{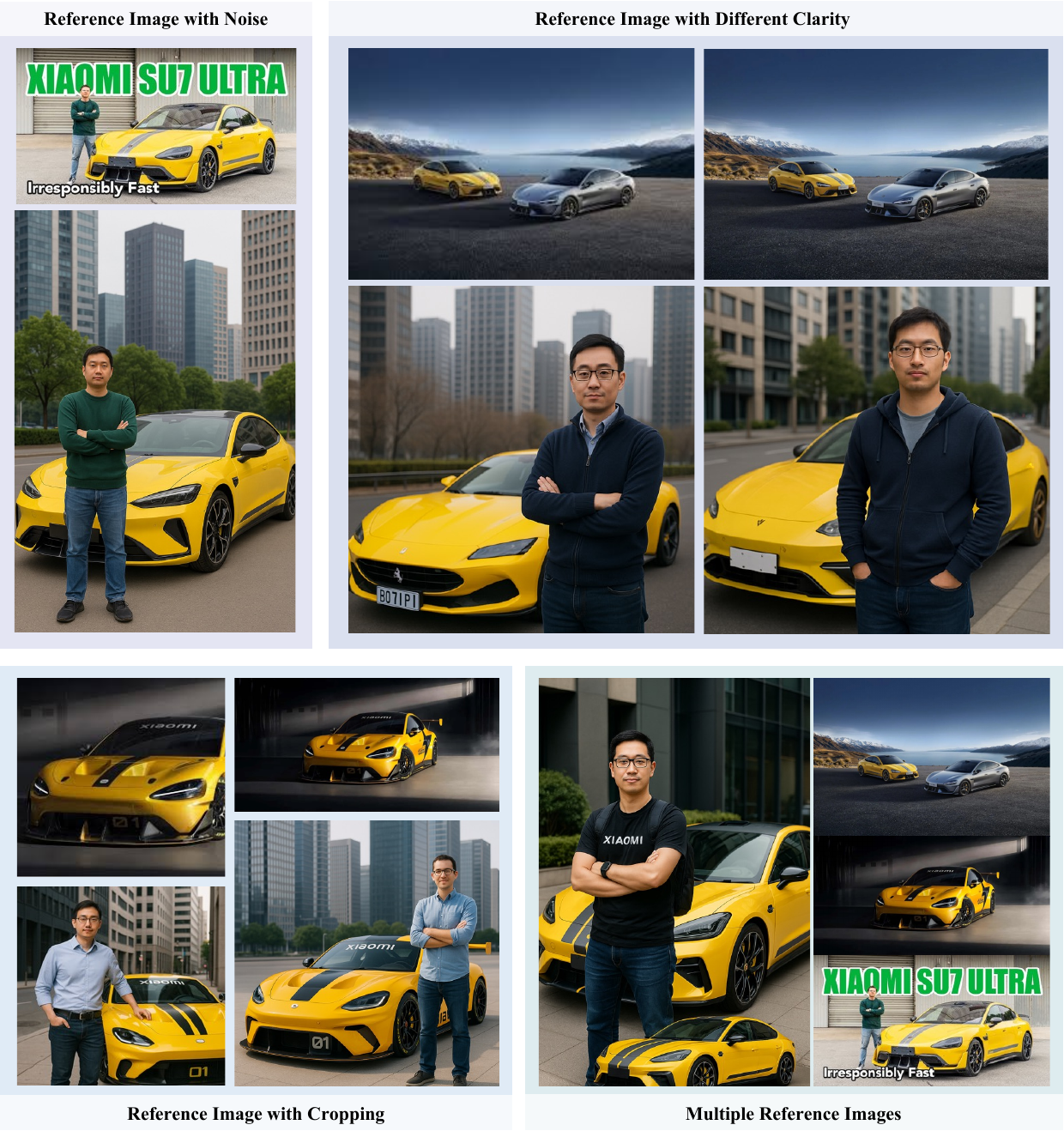}}
\caption{Effect of Retrieved Images, Prompt is "Generate a photo of a typical Xiaomi SU7 Ultra owner and his car. The owner is a 30-40 year old tech professional living in a major city in China, with an relative high annual income. The car should resemble the auto in the reference image <image\#X>. Ignore unrelated details."}
\label{fig:image}
\vspace{-0.3 cm}
\end{figure*}

In this section, we examine how the properties of retrieved reference images affect image generation quality under controlled settings. Using GPT-Image as the generation model, we fix the prompt and accompanying control instructions, and vary four key image conditions: (1) presence of noisy entities, (2) image clarity, (3) cropping, and (4) the number of retrieved images. As illustrated in Figure~\ref{fig:image}, several important observations emerge.

We observe that the quality and structure of retrieved images significantly influence generation outcomes. First, images containing noisy entities, such as unrelated figures or text, can introduce semantic conflicts, leading the model to incorporate incorrect elements. Second, low-resolution or blurry images often cause hallucinations, as the model fills in missing details from prior knowledge, sometimes inaccurately. Third, cropped images that show only partial views (e.g., just the car front) may result in incomplete generations, while full-object images lead to more coherent outputs. Lastly, although multiple images can enrich context, inconsistencies among them can confuse the model and reduce generation fidelity.

\section{Annotators}
We employed a group of trained human annotators through a controlled crowd-sourcing process for factual alignment and multimodal relevance verification in FIG-Eval. All annotators possessed at least TEM-4 or TEM-8 English proficiency and had prior experience in multimodal annotation, such as image–text alignment, caption verification, and factual consistency assessment. Before formal annotation, all participants underwent task-specific training to ensure their understanding of the evaluation criteria and annotation interface.

\textbf{Annotation Guidelines and Criteria.} To ensure annotation reliability and consistency, we prepared a detailed guideline document that provided comprehensive annotation instructions and examples. The guideline covered the following five aspects: (1) entity and concept classification, defining rules for identifying and labeling the ten entity classes and three concept dimensions (Perceptual, Compositional, and Temporal); (2) factual consistency criteria, specifying how to verify whether a generated image accurately reflects real-world facts, attributes, and relations described in the prompt; (3) visual–textual alignment, outlining procedures for assessing semantic and perceptual consistency between textual instructions and image content; (4) error categorization, defining common error types such as hallucinated objects, incorrect attributes, and temporal mismatches; and (5) annotation examples, providing both correct and incorrect examples with justification to standardize annotator judgment.

\textbf{Training and Qualification Phase.} All annotators participated in a trial annotation phase using a shared subset of data, which served both as a training exercise and qualification assessment. Only annotators who demonstrated strong agreement with expert annotations in this phase were retained for the main annotation task. During formal annotation, each generated image–prompt pair was independently reviewed by two annotators and cross-checked for inter-annotator consistency. Any disagreement was resolved by senior annotators with prior experience in factual verification tasks.

\textbf{Quality Control and Review Mechanism.} To further ensure data quality, we implemented a multi-level review mechanism. Automatic filtering removed low-quality or incomplete generations based on resolution and object detection confidence. Human validation was conducted to eliminate NSFW or inappropriate content and to ensure factual and visual coherence. Additionally, non-author experts periodically audited random samples and provided feedback to maintain consistent annotation standards.

\textbf{Ethical Considerations and Compensation.} All annotators were fairly compensated at or above local academic assistant rates. All data used in FIG-Eval were collected from publicly available web sources strictly for academic research. No personally identifiable or sensitive information was included. For the human annotation component, all participants were informed of the research purpose and provided explicit consent prior to participation. Annotators acknowledged that their work would be used solely for dataset construction and model evaluation in research contexts.

\textbf{Use of LLM Assistance in Annotation Support.} In addition, we used ChatGPT as an auxiliary tool to assist in annotation guideline drafting, prompt rephrasing, and preliminary quality checking. ChatGPT was employed solely for non-decisive support tasks such as generating example instructions, verifying language clarity, and identifying potential annotation inconsistencies. All final annotations, factual judgments, and quality evaluations were conducted and verified by trained human annotators. No sensitive data were shared with the model, and all usage strictly adhered to OpenAI’s terms of service and institutional research ethics policies.

\end{document}